\documentclass[journal]{IEEEtran}
\usepackage{cite}
\ifCLASSINFOpdf

\else

\fi

\usepackage{amsfonts,amssymb}
\usepackage{graphicx}
\usepackage{algorithm}
\usepackage{algorithmic}
\usepackage{amsmath}
\usepackage{booktabs}
\usepackage{multirow}
\usepackage{caption}
\usepackage{color}
\usepackage[switch]{lineno}
\usepackage{ragged2e}
\usepackage{paralist}
\usepackage{enumitem}
\usepackage{marginnote}

\usepackage{times}
\usepackage{amsmath}
\usepackage{amssymb}

\usepackage{subfigure}
\usepackage{cite}

\usepackage{url}
\usepackage{amsfonts,amssymb}
\usepackage{graphicx}
\usepackage{algorithm}
\usepackage{algorithmic}
\usepackage{amsmath}
\usepackage{booktabs}
\usepackage{multirow}
\usepackage{caption}
\usepackage{color}
\usepackage[switch]{lineno}
\usepackage{ragged2e}
\usepackage{paralist}
\usepackage{enumitem}
\usepackage{lineno}

\begin{document}
% \linenumbers

%\linenumbers

\title{Weakly Supervised Person Re-ID: Differentiable Graphical Learning and A New Benchmark}

\author{Guangrun Wang, Guangcong Wang, Xujie Zhang, Jianhuang Lai, Zhengtao Yu, and Liang Lin
\IEEEcompsocitemizethanks {\IEEEcompsocthanksitem
The first two authors contribute equally and share the first authorship. G. Wang, G. Wang, X. Zhang, J. Lai and L. Lin are with the School of Data and Computer Science, Sun Yat-sen University, Guangzhou, P. R. China. Email: wanggrun@mail2.sysu.edu.cn.; wanggc3@mail2.sysu.edu.cn; stsljh@mail.sysu.edu.cn; linliang@ieee.org; 
Z. Yu is with Artificial Intelligent Key Laboratory of Yunnan province, Kunming University of Science and Technology, Kunming, 650500, China; Email:ztyu@hotmail.com. Corresponding author: Liang Lin.}}% <-this % stops a space

%\markboth{IEEE TRANSACTIONS ON Pattern Analysis and Machine Intelligence}%
\markboth{IEEE TRANSACTIONS ON Neural Networks and Learning Systems}
{G. Wang\MakeLowercase{\textit{et al.}}: Weakly Supervised Person Re-ID: Differentiable Graphical Learning and A New Benchmark}

\IEEEcompsoctitleabstractindextext{
\begin{abstract}
Person re-identification (Re-ID) benefits greatly from the accurate annotations of existing datasets (e.g., CUHK03 \cite{li2014deepreid} and Market-1501 \cite{zheng2015scalable}), which are quite expensive because each image in these datasets has to be assigned with a proper label. In this work, we ease the annotation of Re-ID by replacing the accurate annotation with inaccurate annotation, i.e., we group the images into bags in terms of time and assign a bag-level label for each bag. This greatly reduces the annotation effort and leads to the creation of a large-scale Re-ID benchmark called SYSU-30$k$. The new benchmark contains $30k$ individuals, which is about $20$ times larger than CUHK03 ($1.3k$ individuals) and Market-1501 ($1.5k$ individuals), and $30$ times larger than ImageNet ($1k$ categories). It sums up to 29,606,918 images. Learning a Re-ID model with bag-level annotation is called the weakly supervised Re-ID problem. To solve this problem, we introduce a differentiable graphical model to capture the dependencies from all images in a bag and generate a reliable pseudo label for each person image. The pseudo label is further used to supervise the learning of the Re-ID model. When compared with the fully supervised Re-ID models, our method achieves state-of-the-art performance on SYSU-30$k$ and other datasets. The code, dataset, and pretrained model will be available at \url{https://github.com/wanggrun/SYSU-30k}. 
\end{abstract}

% Note that keywords are not normally used for peer-review papers.
\begin{IEEEkeywords}
Weakly Supervised Learning, Person Re-identification, Graphical Neural Networks, Visual Surveillance
\end{IEEEkeywords}}

\maketitle

\IEEEdisplaynotcompsoctitleabstractindextext

\IEEEpeerreviewmaketitle

{\section{Introduction}
\label{sect:intro}

\IEEEPARstart{P}{erson} re-identification (Re-ID) \cite{gray2008viewpoint} has been extensively studied in recent years, which refers to the problem of recognizing persons across cameras. Solving the Re-ID problem has many applications in video surveillance for public safety. In the past years, deep learning has been introduced to Re-ID and has achieved promising results.

However, a crucial bottleneck in building deep models is that they typically require strongly annotated images during training. In the context of Re-ID, strong annotation refers to assigning a clear person ID for each person image, which is very expensive because it is difficult for annotators to remember persons who are strangers to the annotators, particularly when the crowd is massive. Moreover, due to the wide range of human activities, many images must be annotated within a rather short time (see Fig. \ref{fig:weakly} (a)).

An alternative way to create a Re-ID benchmark is to replace image-level annotations with bag-level annotations. Suppose that there is a short video containing many person images; we do not need to know who is in each image. A cast of characters is enough. Here, the exact person ID of each image is called the image-level label (Fig. \ref{fig:weakly} (a)), and the cast of characters is called the bag-level label (Fig. \ref{fig:weakly} (b)). Based on our experience, collecting bag-level annotations is approximately three times faster/cheaper than collecting image-level annotations. Once the dataset has been collected, we train Re-ID models with the bag-level annotations. We call this the \emph{weakly supervised Re-ID problem}.

\begin{figure}
\begin{center}
   \includegraphics[width=1.0\linewidth]{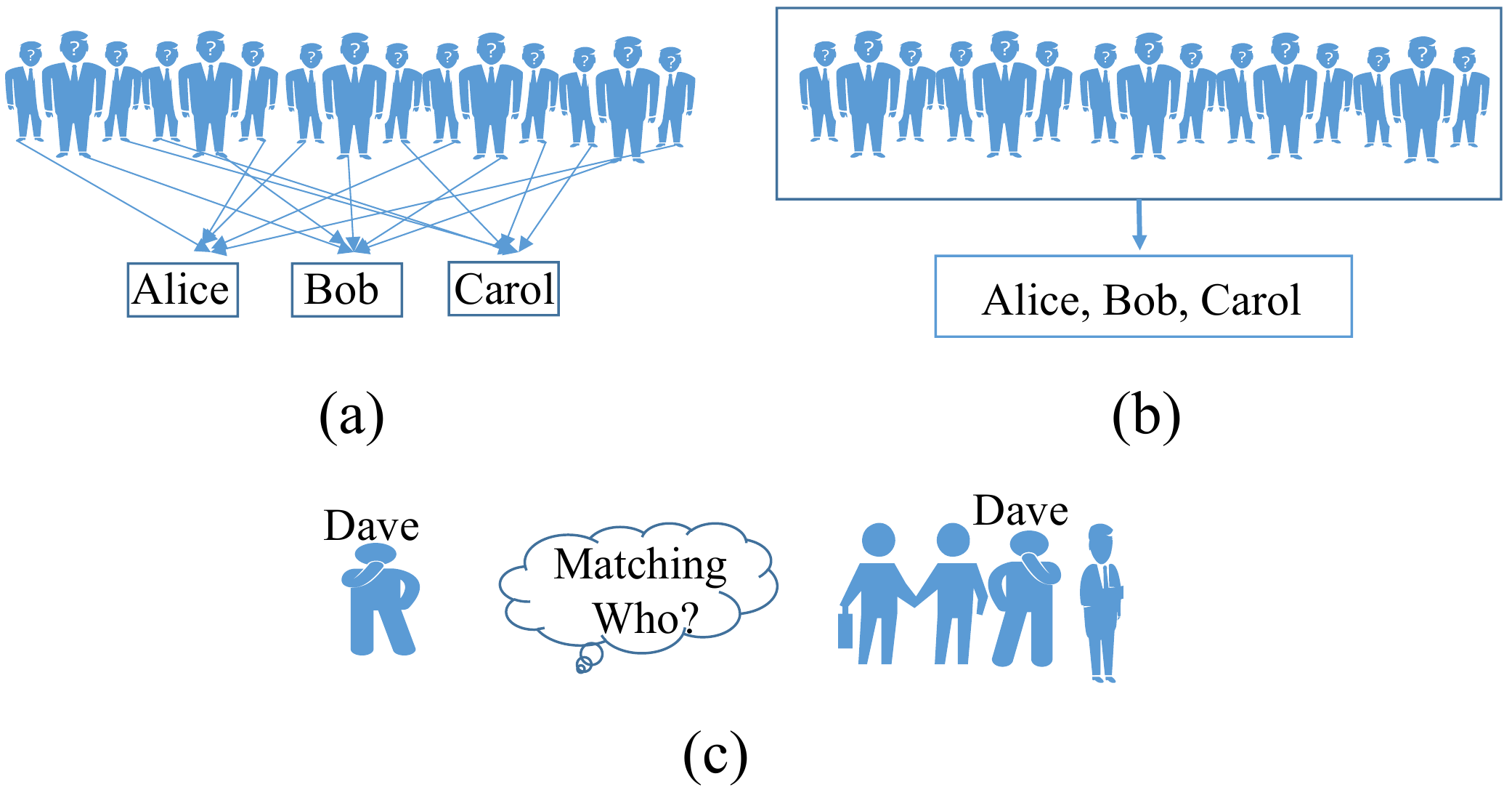}
\end{center}
\vspace{-11pt}
\caption{\small{Problem definition for the weakly supervised Re-ID. (a) is an example of strong annotation while (b) is an example of a weak annotation. During testing, there is no difference between the fully and weakly supervised Re-ID problems, i.e., they both aim at finding the best-matching image for a given person image, as shown in (c).}}\label{fig:weakly}
\vspace{-11pt}
\end{figure}

Formally, with strong supervision, the supervised learning task is to learn $f: X\rightarrow Y$ from a training set
$\{(x_1, y_1), \cdots,(x_i, y_i), \cdots\}$, where $x_i\in X$ is a person image and $y_i\in Y$ is its exact person ID. By contrast, the weakly supervised learning task here is to learn $f: B\rightarrow L$ from a training set $\{(b_1, l_1), \cdots,(b_j, l_j), \cdots\}$, where $b_j\in B$ is a bag of person images, i.e., $b_j = \{x_{j1}, x_{j2}, \cdots, x_{jp}\}$; and $l_j\in L$ is its bag-level label, i.e., $l_j = \{y_{j1}, y_{j2}, \cdots, y_{jq}\}$. Note that the mappings between $\{x_{j1}, x_{j2}, \cdots, x_{jp}\}$ and $\{y_{j1}, y_{j2}, \cdots, y_{jq}\}$ are unknown. During testing, there is no difference between fully and weakly supervised Re-ID problems (see Fig. \ref{fig:weakly} (c)). 

Solving the weakly supervised Re-ID problem is challenging. Because without the help of strongly labeled data, it is rather difficult to model the dramatic variances across camera views, such as the variances in illumination and occlusion conditions, which makes it very challenging to learn a discriminative representation. 
Existing Re-ID approaches cannot solve the weakly supervised Re-ID problem. Regardless of whether they are designed for computing either cross-view-invariant features or distance metrics, the existing models all assume that a strong annotation of each person image is available. This is also reflected in the existing benchmarking Re-ID datasets, most of which consist of a precise person ID for each image. None of them are designed to train a weakly supervised model.

\begin{figure}
\begin{center}
   \includegraphics[width=1\linewidth]{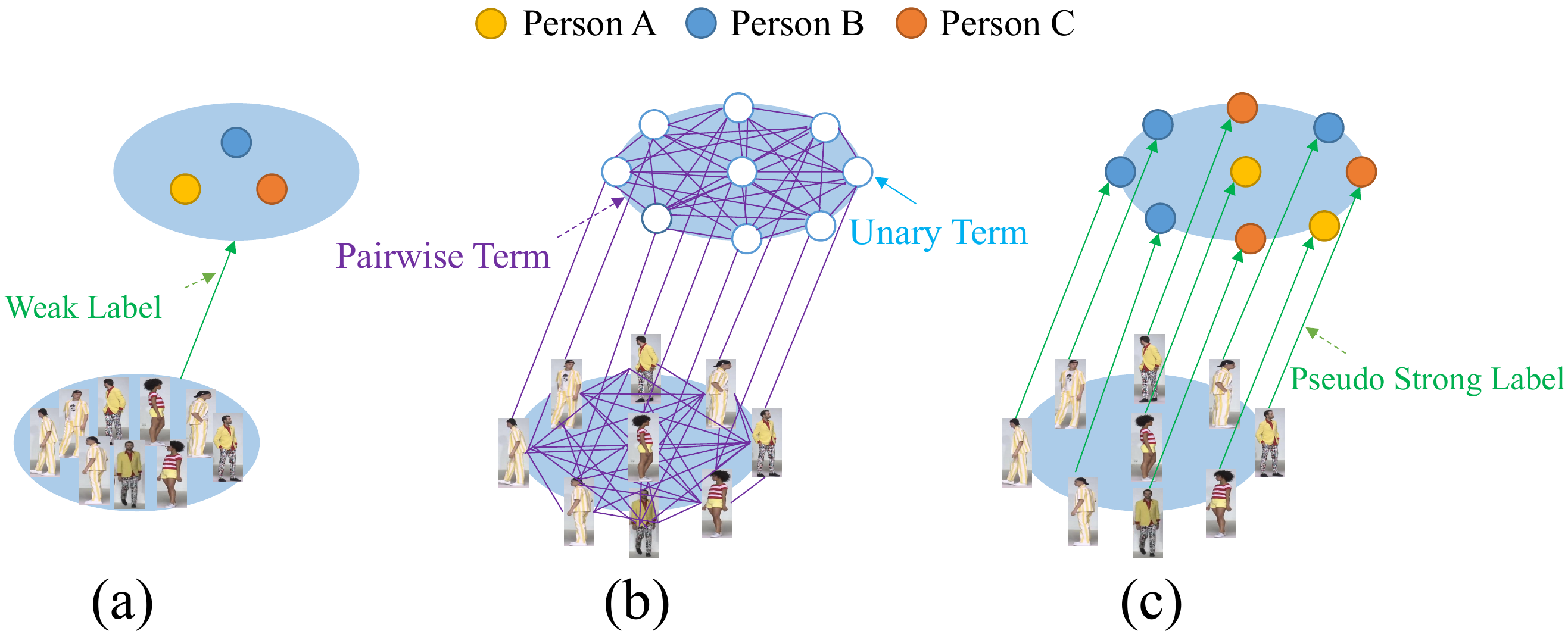}
\end{center}
\vspace{-11pt}
\caption{\small{An illustration of the proposed method for weakly supervised Re-ID. (a) shows a bag of images and their bag-level label.
(b) presents the process of differentiable graphical learning. Using graphical modeling, we can obtain the pseudo image-level label for each image, as shown in (c).}}
\label{fig:crf}
\vspace{-11pt}
\end{figure}

Although the weak annotations lack detailed clues for directly recognizing each person image, they usually contain global dependencies among images, which are very useful to model the variances of images across camera views. By using the weak annotations, we introduce a differentiable graphical model to address the weakly supervised Re-ID problem, which includes several steps. \textbf{First}, the person images are fed into the DNNs in terms of bags (Fig. \ref{fig:crf} (a)) to obtain the rough categorization probabilities. These categorization probabilities are modeled as the unary terms in a probabilistic graphical model; see Fig. \ref{fig:crf} (b). \textbf{Second}, we further model the relations between person images as the pairwise terms in a graph by considering their similarity in the features and appearance; see Fig. \ref{fig:crf} (b). The unary term and the pairwise term are summed to form the refined categorization probability. \textbf{Third}, we maximize the refined categorization probabilities and obtain the pseudo-image-level label for each image. \textbf{Fourth}, we use the generated pseudo labels to supervise the learning of the deep Re-ID model. 
Note that different from traditional non-differentiable graphical models (e.g., CRFs \cite{Lafferty2001Conditional_icml}), our model is differentiable and thus can be integrated into DNNs, which is optimized by using stochastic gradient descent (SGD). All of the above steps are trained in an end-to-end fashion. We summarize the \textbf{contributions} of this work as follows. 

\textbf{1)} We define a weakly supervised Re-ID problem by replacing the image-level annotations in traditional Re-ID with bag-level annotations. This new problem is worth exploring because it significantly reduces the labor of annotation and offers the potential to obtain large-scale training data.

\textbf{2)} Since existing benchmarks largely ignore this weakly supervised Re-ID problem, we contribute a newly dedicated dataset called the SYSU-$30k$ for facilitating further research in Re-ID problems. SYSU-$30k$ contains $30k$ individuals, which is about $20$ times larger than CUHK03 ($1.3k$ individuals) and Market-1501 ($1.5k$ individuals), and $30$ times larger than ImageNet ($1k$ categories). SYSU-$30k$ contains 29,606,918 images. Moreover, SYSU-$30k$ provides not only a large platform for the weakly supervised Re-ID problem but also a more challenging test set that is consistent with the realistic setting for standard evaluation. Fig. \ref{fig:dataset} shows some samples from the SYSU-{{{$30k$}}} dataset.

\begin{figure}
\begin{center}
\vspace{-11pt}
\includegraphics[width=1.0\linewidth]{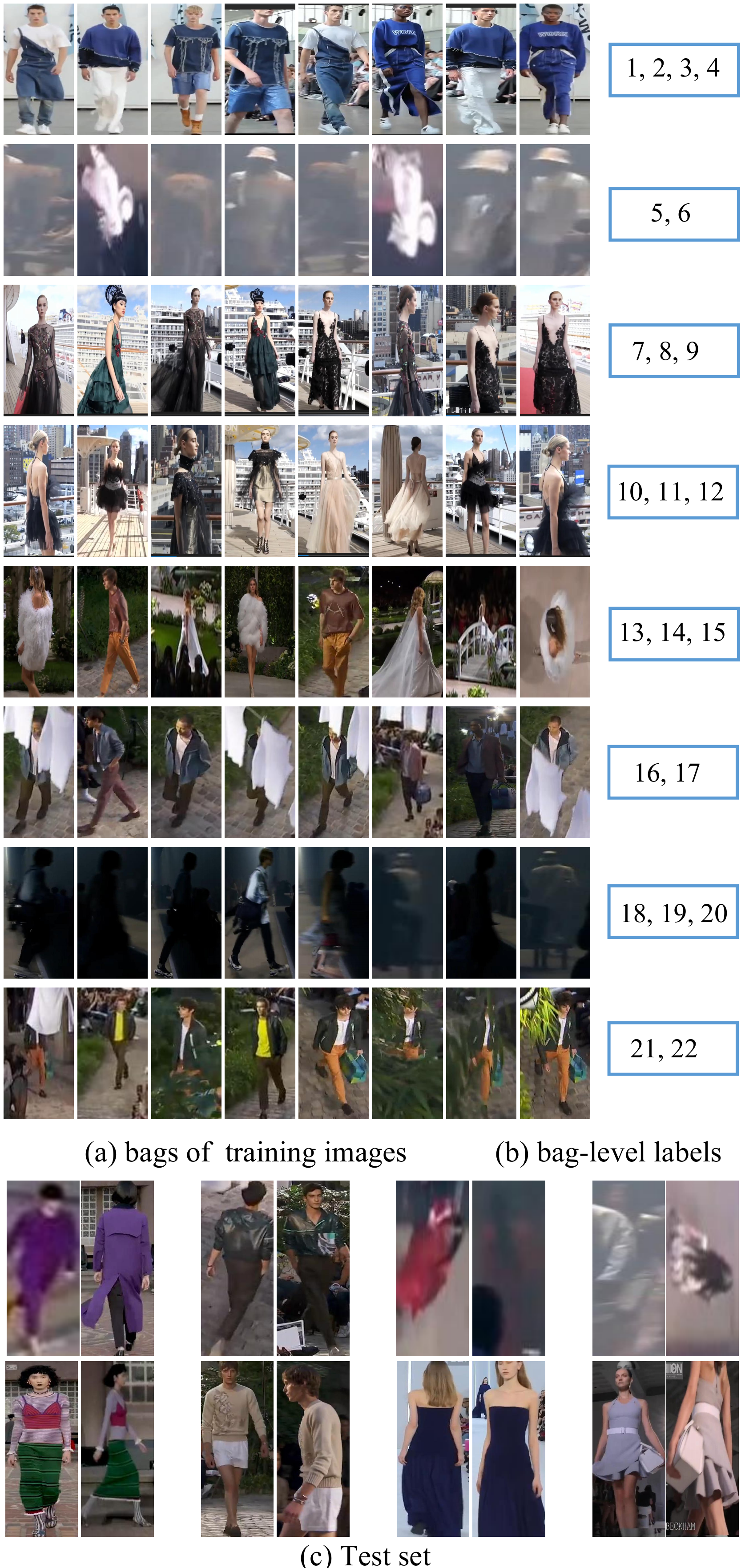}
\end{center}
\vspace{-11pt}
\caption{\small{{Examples in our SYSU-30$k$ dataset. (a) training images in terms of bag; (b) their bag-level annotations; (c) test set.}}}\label{fig:dataset}
\vspace{-22pt}
\end{figure}

\textbf{3)} We introduce a differentiable graphical model to tackle the unreliable annotation dilemma in the weakly supervised Re-ID problem. When compared with the fully supervised Re-ID models, our method achieves state-of-the-art performance on SYSU-30$k$ and other datasets.

The remainder of this work is organized as follows. Section \ref{sect:related_work} provides a brief review of the related work. Section \ref{sect:annotation} introduces the annotation of SYSU-30$k$, followed by the weakly supervised Re-ID model in Section \ref{sect:model}. We also discuss the relationship of our work with previous work in Section \ref{sect:relationship}. The experimental results are presented in Section \ref{sect:exp}. Section \ref{sect:conclusion} concludes the work and presents outlooks for future work.

\vspace{-11pt}
%-------------------------------------------------------------------------
\section{Related Work}\label{sect:related_work}

\textbf{Person Re-ID}. Re-ID has been widely investigated in the literature. Most recent works can be categorized into three groups: (1) extracting invariant and discriminant features \cite{gray2008viewpoint, li2014deepreid}, (2) learning a robust metric or subspace for matching \cite{gray2008viewpoint, koestinger2012large,ding2015deep}, and (3) joint learning of the above two methods \cite{wang2016dari,lin2017cross}. Recently, there are many works on the generalization of Re-ID, such as video-based and image-to-video Re-ID \cite{wang2017p2snet}, spatio-temporal Re-ID \cite{guangcong2019aaai}, occluded Re-ID \cite{zhuo2018occluded}, and natural language Re-ID \cite{li2017person}.
However, all these methods assume that the training labels are strong. They are thus ineffective for solving the weakly supervised Re-ID. Recently, the robustness of Re-ID are also examined by \cite{Wang2020Transferable_cvpr}. 

\textbf{Unsupervised Re-ID.} Another approach that is free from the prohibitively high cost of manual labeling is unsupervised learning Re-ID \cite{yu2017cross,liang2018m2m}. These methods either use local saliency matching or resort to clustering models \cite{yu2017cross}. However, without the help of labeled data, it is difficult to model the dramatic variances across camera views in feature/metric learning. Therefore, it is difficult for these pipelines to achieve high accuracies \cite{song2018unsupervised,yu2019unsupervised,fan2018unsupervised,Li2019Unsupervised_tpami}. In contrast, the proposed weakly supervised Re-ID problem has a better solution. Note that compared to unsupervised Re-ID, the annotation effort of weakly supervised Re-ID is also very inexpensive.

\textbf{Semi-supervised Re-ID.} Apart from our model, there have been some uncertain label learning models, among which the one-shot/one-example Re-ID\cite{wu2019progressive,wu2018exploit} is the most related to ours. The main differences between their methods and ours are two-fold. First, in one-shot Re-ID, at least one accurate label for each person ID is still in desire. While in our weakly supervised Re-ID, no accurate label is needed. Second, there are bag-level labels as constraints to guide the estimation of the pseudo labels in our method, ensuring that our generated pseudo labels to be more reliable than those generated by one-shot Re-ID. Besides, \cite{meng2019weakly} also proposes to cope with the uncertain-label Re-ID problem using multiple-instance multiple-label learning. However, similar to \cite{wu2018exploit}, at least one accurate label for each person ID is still in a desire to form the probe set in \cite{meng2019weakly}. Therefore, mathematically, \cite{wu2019progressive,wu2018exploit,meng2019weakly} are all semi-supervised Re-ID but NOT weakly supervised Re-ID.

\textbf{Weakly-supervised learning.} Beyond Re-ID, although training deep models with weak annotations is challenging, it has been partially investigated in the literature, such as in tasks of image classification \cite{mahajan2018exploring,wang2017deep}, semantic segmentation \cite{wang2017learning,luo2017deep,lin2016deep,Zhang2019Hierarchical_pami}, object detection \cite{cinbis2017weakly}. Our method is related to them in that our model is also based on the generation of a pseudo label. However, the weakly supervised Re-ID problem has two unique characteristics that distinguish it from other weakly supervised learning tasks. (1) We cannot find a representative image for a permanent ID because people will change their clothes at short intervals. The same person wearing different clothes may be regarded as two different persons. This results in thousands of millions of person IDs. Therefore, the label for a weakly supervised Re-ID sample is fuzzier than other tasks. (2) The entropy of the weakly supervised Re-ID problem is larger than other tasks. In weakly supervised segmentation tasks, pixels in an image share certain motion of rigidity and stability that benefits the prediction. Whereas in the weakly supervised Re-ID, persons in video bags are more unordered and irregular. Therefore, the weakly supervised Re-ID problem is considerably more challenging than other problems.

\textbf{Graphical learning.} To address the weakly supervised Re-ID problem, we propose to generate a pseudo label for each image by introducing differentiable graphical learning, which is inspired by the advances in image segmentation \cite{chen2018deeplab} and videos \cite{wu2013constrained}. Recently, one classical graphical model, i.e., CRF, has been introduced to Re-ID for similarity learning\cite{shen2018person}. However, our method differs from \cite{shen2018person} in two aspects. First, like existing methods, \cite{shen2018person} uses CRF as a post-processing tool to refine the prediction provided by fully supervised learning, while our method exploits the supervision-independent property of graphical learning \cite{chen2018deeplab} to generate pseudo labels for our weakly-supervised Re-ID learning. Second, different from traditional non-differentiable graphical models and \cite{shen2018person}, our proposed model directly formulates the graphical learning as an additional loss, which is differentiable to the neural network parameters and thus can be optimized by using stochastic gradient descent (SGD).

\begin{table*}[t]
\vspace{-11pt}
\caption{\small{A comparison of different Re-ID benchmarks. \textbf{Categories:} We treat each person identity as a category. \textbf{Scene:} whether the video is taken indoors or outdoors. \textbf{Annotation:} whether image-level labels are provided. \textbf{Images:} the person images which are obtained by using a human detector to detect the video frames. Actually, the person images in this work refer to the bounding boxes.}}\label{tab:dataset_cmp}
\centering
\scriptsize
\begin{tabular}{|c|c|c|c|c|c|c|c|c|c|}
\multicolumn{10}{c}{\small{(a) Comparision with existing Re-ID datasets.}} \\ 
\hline
Dataset & CUHK03\cite{li2014deepreid} & Market-1501\cite{zheng2015scalable} & Duke\cite{zheng2017unlabeled} & MSMT17\cite{wei2018person} & CUHK01\cite{li2012human} & PRID\cite{hirzer2011person} & VIPeR\cite{gray2008viewpoint} & CAVIAR\cite{cheng2011custom} & SYSU-$30k$\\ \hline
Categories & 1,467 & 1,501 & 1,812 & 4,101 & 971 & 934 & 632 & 72 & \textbf{30,508} \\ \hline
Scene & Indoor & Outdoor & Outdoor & Indoor, Outdoor & Indoor & Outdoor & Outdoor & Indoor & \textbf{Indoor, Outdoor} \\ \hline
Annotation & Strong & Strong & Strong & Strong & Strong & Strong & Strong & Strong & \textbf{Weak} \\ \hline
Cameras & 2 & 6 & 8 & 15 & 10 & 2 & 2 & 2 & \textbf{Countless} \\ \hline
Images & 28,192 & 32,668 & 36,411 & 126,441 & 3,884 & 1,134 & 1,264 & 610 & \textbf{29,606,918} \\ \hline
\multicolumn{10}{c}{} \\ 
\multicolumn{10}{c}{\small{(b) Comparison with ImageNet-1$k$}}\\ 
\hline
\multicolumn{2}{|c|}{Dataset} & \multicolumn{4}{c|}{ImageNet-1$k$} & \multicolumn{4}{c|}{SYSU-$30k$} \\ \hline
\multicolumn{2}{|c|}{Categories} & \multicolumn{4}{c|}{1,000} & \multicolumn{4}{c|}{\textbf{30,508}} \\ \hline
\multicolumn{2}{|c|}{Images} & \multicolumn{4}{c|}{1,280,000} & \multicolumn{4}{c|}{\textbf{29,606,918}} \\ \hline
\multicolumn{2}{|c|}{Annotation} & \multicolumn{4}{c|}{Strong} & \multicolumn{4}{c|}{\textbf{Weak}} \\ \hline
\end{tabular}
\vspace{-11pt}
\end{table*}

\textbf{Person search.} Another problem that is very related to our problem is person search \cite{li2017person}, which aims to
fuse the processes of person detection and Re-ID.
There are two significant differences between weakly supervised Re-ID and person search.
First, the weakly supervised Re-ID only focuses on visual matching, which is reasonable because current human detectors are competent enough to detect persons.
Second, the weakly supervised Re-ID problem enjoys the inexpensive efforts of weak annotation, while the person search still needs a strong annotation for each person image.

\section{SYSU-{{{$30k$}}} Dataset} \label{sect:annotation}
\textbf{Data collection.} No weakly supervised Re-ID dataset is publicly available. To fill this gap, we contribute a new Re-ID dataset named SYSU-{$30k$} in the wild to facilitate studies. We download many short program videos from the Internet. TV programs are considered as our video source for two reasons. \textbf{First}, the pedestrians in a TV program video are often cross-view and cross-camera because there are many movable cameras to capture the shots for post-processing. Re-identifying pedestrians in a TV program video is exactly a Re-ID problem in the wild. \textbf{Second}, the number of pedestrians in a program is suitable for annotation. On average, each video contains 30.5 pedestrians walking around.

Our final raw video set contains 1,000 videos. The annotators are then asked to annotate the videos in a weak fashion. In particular, each video is divided into 84,924 bags of arbitrary length. Then, the annotators record the pedestrian's identity for each bag. YOLO-v2 \cite{redmon2017yolo9000} is utilized for pedestrian bounding box detection. Three annotators review the detected bounding boxes and annotate person category labels for 20 days. Finally, 29,606,918 ($\approx30M$) bounding boxes of 30,508 ($\approx30k$) person categories are annotated. We then select 2,198 identities as the test set, leaving the rest as the training set. There is no overlap between the training set and the test set. Notably, we treat each person identity as a category in this paper.

{We provide some samples of the SYSU-{{{$30k$}}} dataset in Fig. \ref{fig:dataset}. As shown, our SYSU-30$k$ dataset exhibits challenges of illumination variance (Row 2, 7, and 9), occlusion (Row 6 and 8), low resolution (Row 2 and 9), looking-downward cameras (Row 2, 5, 6, 8, and 9), and complicated backgrounds of the real scenes (Row 2-10).}

\textbf{Dataset statistics.} SYSU-{{{$30k$}}} contains 29,606,918 person images with 30,508 categories in total, which is further divided into 84,930 bags (only for training set). Fig. \ref{fig:dataset_statistis} (a) summarizes the number of bags with respect to the number of images per bag, showing that each bag has 2,885 images on average. This histogram reveals the person image distribution of these bags in the real world without any manual cleaning and refinement.
Each bag is provided with an annotation of bag-level labels. 

\textbf{Comparison with existing Re-ID benchmarks.} We compare SYSU-$30k$ with existing Re-ID datasets, including CUHK03 \cite{li2014deepreid}, Market-1501 \cite{zheng2015scalable}, Duke \cite{zheng2017unlabeled}, MSMT17 \cite{wei2018person}, CUHK01 \cite{li2012human}, PRID \cite{hirzer2011person}, VIPeR \cite{gray2008viewpoint}, and CAVIAR \cite{cheng2011custom}. Fig. \ref{fig:dataset_statistis} (c) and (d) plots the person IDs and the number of images, respectively, indicating that SYSU-$30k$ is much larger than existing datasets.
To evaluate the performance of the weakly supervised Re-ID approach, we randomly choose 2,198 person categories from SYSU-$30k$ as the test set. These person categories are not utilized in training. We annotate an accurate person ID for each person image. We also compare the test set of SYSU-$30k$ with existing Re-ID datasets. From Fig. \ref{fig:dataset_statistis} (b) and (c), we can observe that the test set of SYSU-$30k$ is more challenging than those of the competitors in terms of both the image number and person IDs. Thanks to the above annotation fashion, the SYSU-{{{$30k$}}} test set can adequately reflect the real-world setting and is consequently more challenging than existing Re-ID datasets. Therefore, SYSU-{{{$30k$}}} is not only a large benchmark for the weakly supervised Re-ID problem but is also a significant standard platform for evaluating existing fully-supervised Re-ID methods in the wild.

A further comparison of SYSU-$30k$ with existing Re-ID benchmarks is shown in Table \ref{tab:dataset_cmp} (a), including categories, scene, annotation, cameras, and image numbers (bounding boxes). After the comparison, we summarize the new features in SYSU-{$30k$} in the following aspects. First, SYSU-{$30k$} is the first weakly annotated dataset for Re-ID. Second, SYSU-$30k$ is the largest Re-ID dataset in terms of both person categories and image number. Third, SYSU-$30k$ is more challenging due to many cameras, realistic indoor and outdoor scenes, and occasionally incorrect annotations. Fourth, the test set of SYSU-$30k$ is not only suitable for the weakly supervised Re-ID problem but is also a significant standard platform to evaluate existing fully supervised Re-ID methods in the wild.

\begin{figure}
 \begin{center}
\includegraphics[width=0.67\linewidth]{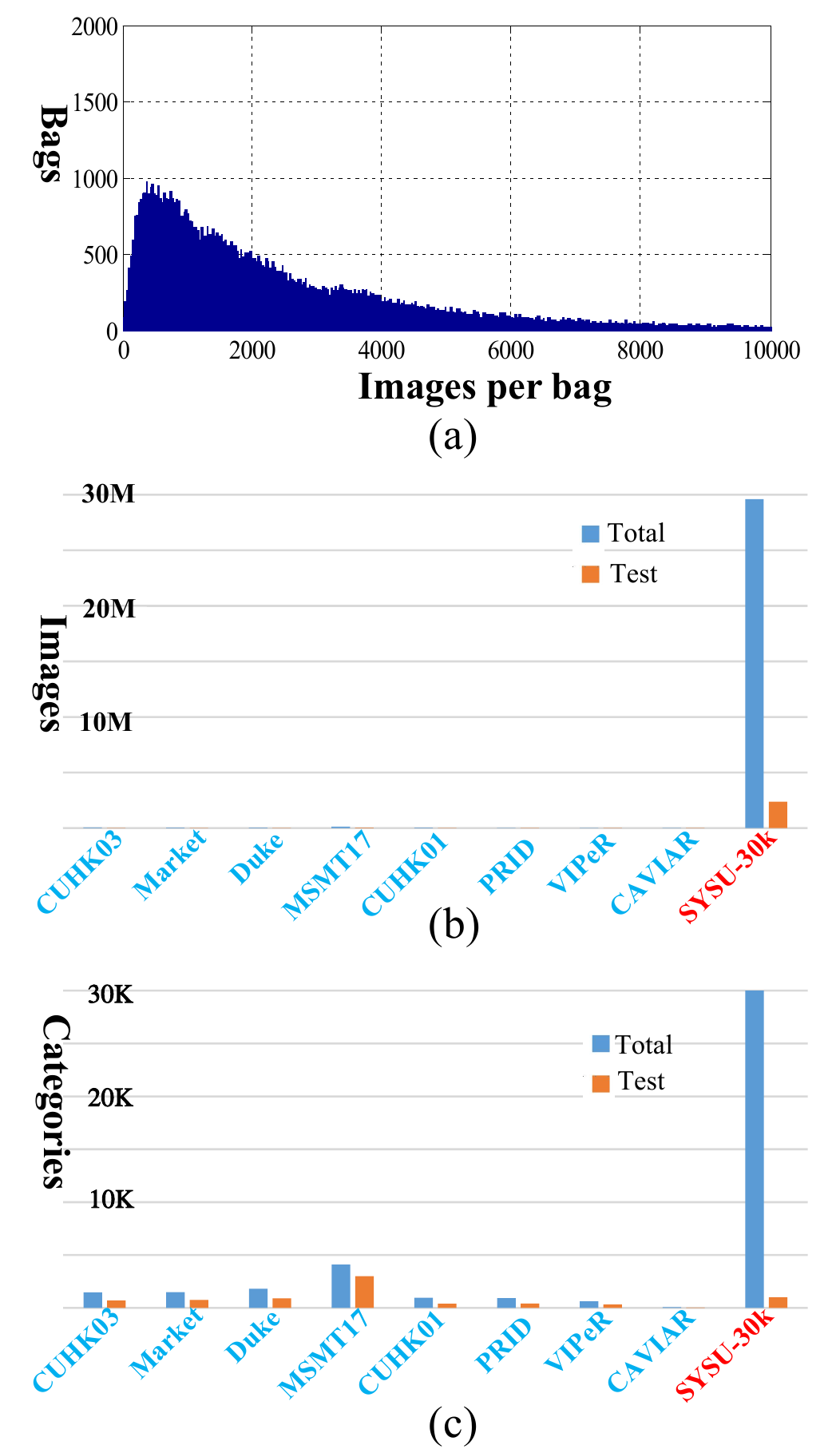}
\end{center}
\vspace{-11pt}
\caption{\small{The statistics of the SYSU-{{{$30k$}}}. (a) summarizes the number of the bags with respect to the number of the images per bag. (b) and (c) compare SYSU-$30k$ with the existing datasets in terms of image number and person IDs for both the entire dataset and the test set.}}\label{fig:dataset_statistis}
\vspace{-11pt}
\end{figure}

\textbf{Comparison with ImageNet-1$k$.} Beyond the Re-ID family, we also compare SYSU-$30k$ with the well-known ImageNet-1$k$ benchmark for general image recognition. As shown in Table \ref{tab:dataset_cmp} (b), SYSU-$30k$ has several appealing advantages over ImageNet-1$k$. First, SYSU-$30k$ has more object categories than ImageNet-1$k$, i.e., 30$k$ vs 1$k$. Second, SYSU-$30k$ saves annotation due to the effective weak annotation.

\textbf{Evaluation protocol.} The evaluation protocol of SYSU-$30k$ is similar to that of the previous datasets \cite{zheng2015scalable}. Following \cite{li2014deepreid}, we fix the train/test partitioning. In the test set, we choose 1,000 images belonging to 1,000 different person IDs to form the query set. As the scalability is important for the practicability of Re-ID systems, we propose to challenge the scalability of a Re-ID model by providing a gallery set containing a vast volume of distractors for validation. Specifically, for each probe, there is only one matching person image as the correct answer in the gallery, while there are 478,730 mismatching person images as the wrong answer in the gallery. Thus, the evaluation protocol is to search for a needle in the ocean, just like the police search a massive amount of videos for a criminal. Following \cite{zheng2015scalable}, we use the rank-1 accuracy as the evaluation metric.  

\vspace{-2pt}
\section{Weakly Supervised Re-ID Model}\label{sect:model}
 
\subsection{From Supervised Re-ID to Weakly Supervised Re-ID}\label{sect:full}
{Let $b= \{x_{1}, \cdots, x_{j}, \cdots, x_{p}\}$ denote a bag containing $p$ images. $y= \{y_{1}, \cdots, y_{j}, \cdots, y_{p}\}$ are the image-level labels and $l$ is the bag-level label. }

{In the fully supervised Re-ID, the image-level labels $y$ are known. The goal is to learn a model by minimizing the loss between the image-level labels and the predictions.}

{In contrast, in the weakly supervised Re-ID, the bag-level label $l$ is provided, but the image-level labels $y$ are unknown. Suppose a bag contains $n$ person IDs and there are $m$ IDs in the entire dataset. A preliminary image-level label $\mathbf{Y}_{j}$ for each image $x_{j}$ can be inferred from its bag-level label $l$:\begin{scriptsize}\begin{equation}\label{eq:prob}
{\mathbf{Y}}_{j} = \left(\begin{array}{c} {\mathbf{Y}}_{j}^{1} \\\vdots \\ {\mathbf{Y}}_{j}^{k} \\ \vdots \\ {\mathbf{Y}}_{j}^{m}\end{array}\right), where ~~~{\mathbf{Y}}_{j}^{k} = \Bigg\{\begin{array}{c} \frac{1}{n}, ~~~if ~~~k \in l\\ {} \\0, ~~~ otherwise~~~\end{array},
\end{equation}\end{scriptsize}Then, $\mathbf{Y}$ can be used as a bag constraint to deduce a pseudo-image-level labels $\hat{y}$, which can be further used to supervise the model learning.}

\subsection{Weakly supervised Re-ID: differentiable graphical learning}\label{sect:crfasdnn}%{sect:weak2full}

{In this section, we propose differentiable graphical learning to generate pseudo-image-level labels for the person images.}

\vspace{5pt}
{\textbf{Graphically modeling Re-ID}. In our graph, each node represents a person image $x_i$ in a bag and each edge represents the relation between person images, as illustrated in Fig. \ref{fig:graph}. Here $i$ is the image index in a bag. Assigning a label $y_i$ to a node $x_i$ has an energy cost. The energy cost $E(y|x)$ of our graph is defined as:\begin{small}\begin{equation}
\label{eq:crf}
E(y|x) = \underbrace{\sum_{\forall i \in U} \Phi(y_{i}|x_i)}_{\text{unary term}} + \underbrace{\sum_{\forall i, j \in V} \Psi(y_{i}, y_{j}|x_i;x_j)}_{\text{pairwise term}},
\end{equation}\end{small}where $U$ and $V$ denote a set of nodes and edges, respectively. $\Phi(y_{i}|x_i)$ is the unary term measuring the cost of assigning label $y_i$ to a person image $x_i$. $\Psi(y_{i}, y_{j}|x_i;x_j)$ is the pairwise term that measures the penalty of assigning labels to a pair of images $(x_i, x_i)$. Mathematically, graphical modeling is to smooth the uncertain prediction of person IDs. The unary term performs the prediction based on sole nodes. While the pairwise term smoothes the prediction of multiple nodes by considering their appearance and features. In summary, Eq. \eqref{eq:crf} is to clean up the spurious predictions of classifiers learned in a weakly supervised manner.}

\begin{figure}[t]
 \begin{center}
\includegraphics[width=0.95\linewidth]{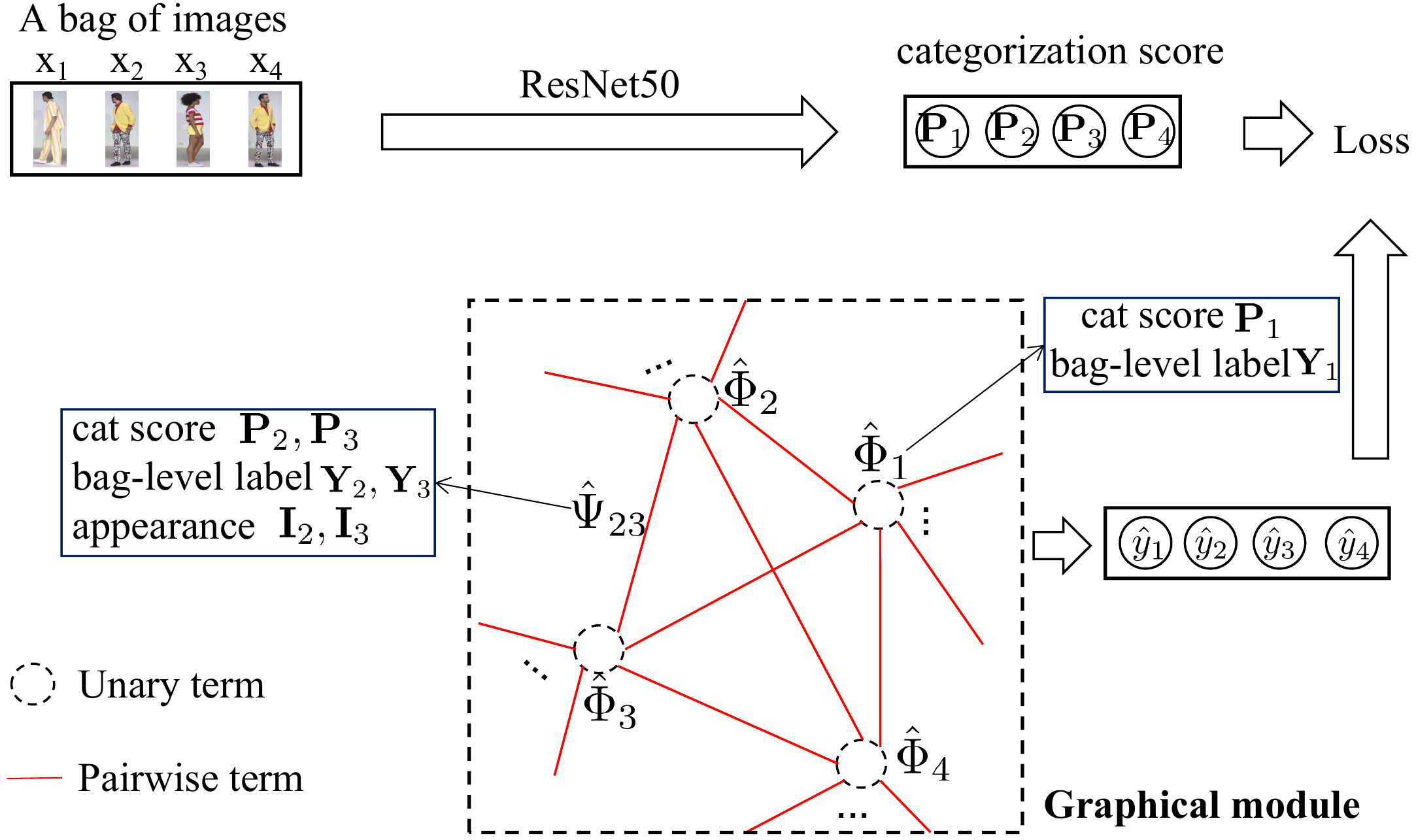}
\end{center}
\vspace{-11pt}
\caption{\small{{Graphical model to generate pseudo image-level labels for person images. The unary terms are estimated by the deep networks, while the pairwise terms involve the similarity of features, the image appearance, and the bag-level label.}}}\label{fig:graph}
\vspace{-11pt}
\end{figure}

{\textbf{Unary term.} The unary terms is typically defined as:\begin{small}\begin{equation}
\label{eq:unary}
\Phi(y_{i}|x_i) = -  \mathbf{P}_i^{y_i}\log ( {\mathbf{Y}}_i^{y_i} \odot \mathbf{P}_i^{y_i} ),
\end{equation}\end{small}where $\mathbf{P}_i$ is the categorization probability of a person image $x_i$ outputted by a DNN. $\odot$ denotes element-wise product.}

{As the unary term alone is generally noisy and inconsistent. Interactions between pairwise terms are required.} 

{\textbf{Pairwise term.} The pairwise term is defined as:\begin{small}
\begin{equation}
\label{eq:pairwise}
\begin{aligned}
\Psi(y_{i}, y_{j}|x_i;x_j) & = \underbrace{\zeta (y_i, y_j)}_{\text{{label compatibility}}} \underbrace{\mathbf{Y}_{i}^{y_i} \mathbf{Y}_{j}^{y_j}}_{\text{{bag constraint}}} \underbrace{\exp \bigg( - \frac{\|I_i - I_j\|^{2}}{2 \sigma^{2}} \bigg)}_{\text{{appearance similarity}}},\\
\end{aligned}
\end{equation}
\end{small}where $I_i$ and $I_j$ are the low-level features and based on them, a Gaussian kernel is employed to measure their appearance similarity. The hyper-parameter $\sigma$ controls the scale of the Gaussian kernel. This kernel forces the images with similar appearance to have the same labels. }
{Similar to the unary term, the pairwise terms are also bounded by the bag-level annotations ${\mathbf{Y}}_{i}$ and ${\mathbf{Y}}_{j}$. The pairwise terms are widely known to provide nontrivial knowledge (e.g., structural context dependencies) that is not captured by the unary term. A simple label compatibility function $\zeta(y_i, y_j) \in \{0,1\}$ in Eq. \eqref{eq:pairwise} is given by the Potts model, namely,
\begin{small}\begin{equation}\label{eq:label_comp}
\zeta(y_i, y_j) = \Bigg\{\begin{array}{c} 0, \text{if    } y_i = y_j\\ {} \\1, \text{otherwise   }\end{array},
\end{equation}\end{small}It introduces a penalty for similar images that are assigned different labels. Considering that Eq. \eqref{eq:crf} is non-differentiable, it is incompatible with DNNs. Thus, we will instead learn a differential version of Eq. \eqref{eq:crf} in a deep learning model.}

{\textbf{Bag constraint.} As mentioned above, both the unary and pairwise terms are constrained by the bag-level annotations ${\mathbf{Y}}_{i}$ and ${\mathbf{Y}}_{j}$. In fact, the bag-level annotation contains extra knowledge that helps to improve the estimation. For example, if the estimator mismatches a person image to an ID that is not in the bag-level annotation, the estimation is undoubtedly considered as incorrect. Then, the estimation will be corrected by matching the image to the ID in the bag-level annotation with the most significant prediction score. Furthermore, if some IDs in the weak annotation are absent in the prediction, the proposed method will encourage a portion of the person images to be assigned to such IDs to improve the performance. In this way, knowledge of the weakly labeled data can be fully exploited. Given a bag of images and their bag-level label, we refine the DNN predictions by element-wisely multiplying $\mathbf{P}$ with the bag-level weak annotation ${\mathbf{Y}}$. This is shown in the unary term in Eq. \eqref{eq:unary}. Similarly, we also impose ${\mathbf{Y}}_i$ and ${\mathbf{Y}}_j$ on the pairwise term in Eq. \eqref{eq:pairwise}.}

{Moreover, there is a natural smoothness in a video that could not be ignored. The person IDs in the adjacent bags change slowly within a short time; for instance, an image-level label $y_i$ in bag $b_T$ could also be in the bag $b_{T+1}$. A large amount of bags with overlapping IDs naturally exist in a video, which sheds light on the ability of the weakly supervised Re-ID. As a special example, if $b_T$ contains $\{y_i,y_j\}$ and $b_{T+1}$ contains $\{y_j,y_k\}$, then the two bags share $\{y_j\}$. Meanwhile, one image in the first bag is similar to another image in the second bag. Hence, our model predicts these two images as $y_i$. Finally, $y_i$ and $y_k$ are assigned to the remaining images.}

\begin{figure}[t]
 \begin{center}
\includegraphics[width=1.0\linewidth]{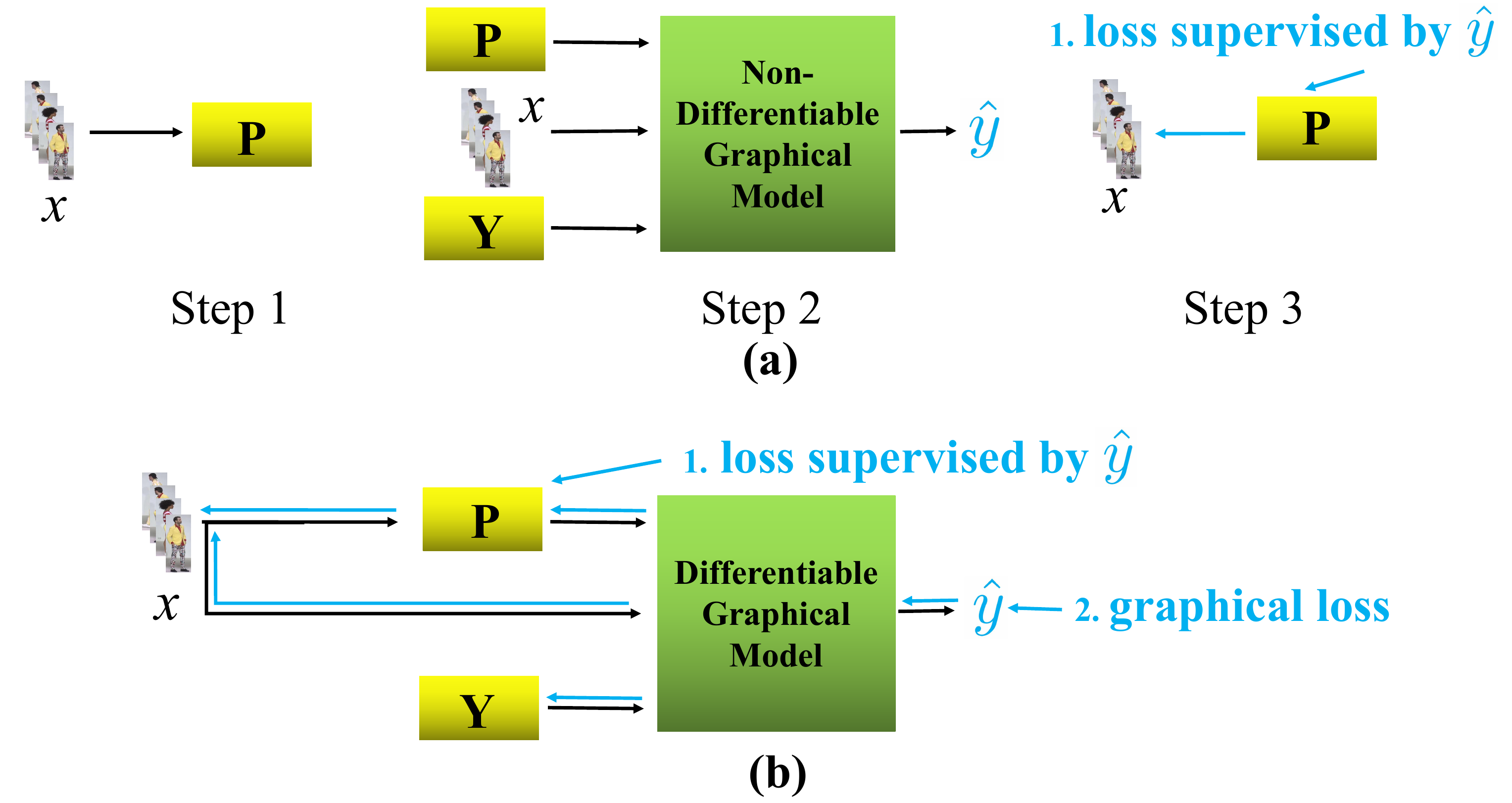}
\end{center}
\vspace{-11pt}
\caption{\small{{Differentiable graphical model in DNNs, where $x, \mathbf{Y}, \mathbf{P}, \hat{y}$ denote the input images, bag-level label, preliminary categorization, and refined categorization, respectively. (a) is the stepwise graphical model, while (b) is our proposed end-to-end differentiable graphical model. Our model consists of two losses, i.e., an unsupervised loss for pseudo label generation and a loss supervised by the pseudo labels. Here, the black lines denote forward passing, while the blue lines denote back-propagation. }}}\label{fig:crfasdnn}
\vspace{-11pt}
\end{figure}

{\textbf{Deducing pseudo image-level labels.} By {minimizing} the energy cost of Eq. \eqref{eq:crf}, we can obtain the pseudo image-level label $\hat{y}_i$ for the person image $\hat{x}_i$:\begin{small}\begin{equation}
\label{eq:pseudo}
\hat{y}_i = \mathop{\arg\max}\limits_{y_i \in \{1,\cdots, m\}}  E(y_i|x_i),
\end{equation}\end{small}where $\{1,\cdots, m\}$ denote all the person IDs in the training set. Once such labels are generated, they are used to update the network parameters.}

\begin{figure*}[t]
\vspace{-11pt}
\begin{center}
   \includegraphics[width=0.72\linewidth]{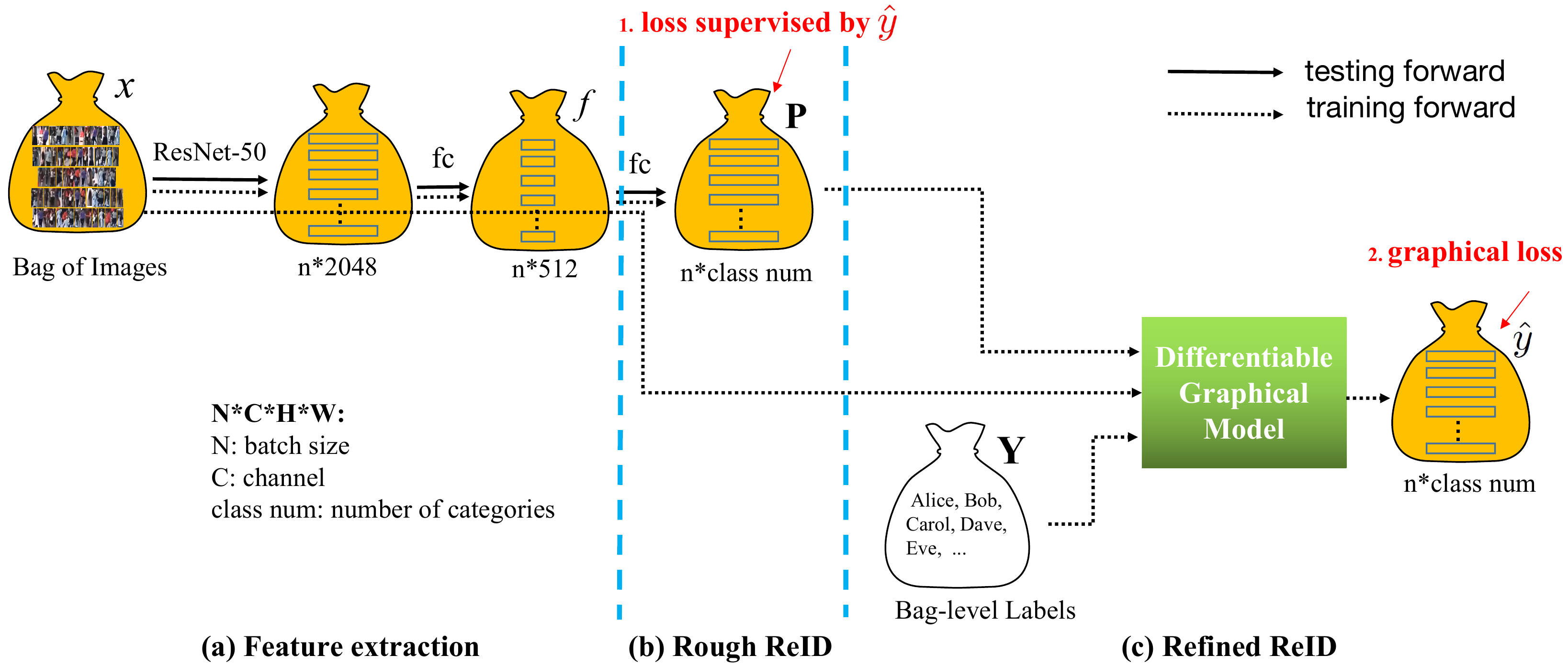}
\end{center}
\vspace{-11pt}
\caption{\small{Diagram of our approach. It mainly consists of three stages, i.e., feature extraction, rough Re-ID, and refined Re-ID. The solid black flow denotes the testing stage, while the black dotted flow denotes the training stage. For simplification, the back-propagation flow is omitted. The loss function is marked with a red arrow.}}\label{fig:method}
\vspace{-11pt}
\end{figure*}

\textbf{Differentiablizing graphical learning.} The above weakly supervised Re-ID model is not end-to-end. Because we must first use an external graphical learning solver to obtain the pseudo labels and then use another solver to train the DNNs under the supervision of the pseudo labels (see Fig. \ref{fig:crfasdnn} (a)). To enable an end-to-end optimization, we propose to make our graphical learning differentiable and compatible with DNNs (see Fig. \ref{fig:crfasdnn} (b)).

{We first investigate the mechanism of a non-differentiable graphical model. As illustrated in Fig. \ref{fig:crfasdnn} (a), a non-differentiable graphical model consists of three steps. \textbf{First}, a preliminary categorization score $\mathbf{P}$ is obtained through a DNN. \textbf{Second}, the energy cost in Eq. \eqref{eq:crf} is minimized by re-assigning labels to the images appropriately, subject to the appearance similarity, the preliminary categorization scores, and the bag constraint. \textbf{Third}, the re-assigned labels are considered as the pseudo labels and used to supervise the learning of the Re-ID model.}

{The label reassignment in the second step claimed above is non-differentiable, which makes the graphical model incompatible with DNNs. To fill this gap, a relaxation form of Eq. \eqref{eq:crf} is desirable. 
With a continuous version of $\hat{\Phi}$ and $\hat{\Psi}$ to approximate the discrete $\Phi$ and $\Psi$, we rewrite Eq. \eqref{eq:crf} as follows 
\begin{small}\begin{equation}\label{eq:crfloss}
\mathcal{L}_{graph}(x) = \underbrace{\sum_{\forall i \in U} \hat{\Phi} (x_{i})}_{\text{unary term}} + \underbrace{\sum_{\forall i, j \in V} \hat{\Psi}(x_{i}, x_{j})}_{\text{pairwise term}},
\end{equation}\end{small}
where $\hat{\Phi}$ and $\hat{\Psi}$ are defined as:\begin{small}\begin{equation}\label{eq:unary-loss}
\left\{
\begin{aligned}
\hat{\Phi}(x_i) &= -\sum\limits_{j=1}^m [h(\mathop{\arg\max}_{k\in\{1,\cdots,m\}}\mathbf{Y}_i^k\odot\mathbf{P}_i^k)]^j\log(\mathbf{P}_i^j),\\
\hat{\Psi}(x_i,x_j) &=  -\exp \bigg( - \frac{\|I_i - I_j\|^{2}}{2 \sigma^{2}} \bigg)(\mathbf{Y}_i \mathbf{P}_i)^{\mathbf{T}} \log(\mathbf{Y}_j \mathbf{P}_j).
\end{aligned}
\right.
\end{equation}\end{small}$\arg\max$ returns the index of the largest element in a vector, $h$ is a function that maps a scalar to one-hot vector, and the superscripts $j$ and $k$ mean indexing the $j$ and $k$ element of a vector, respectively. The differences between Eq. \eqref{eq:unary}-\eqref{eq:pairwise} and Eq. \eqref{eq:unary-loss} are summarized as follows: 
\textbf{1)} 
the replacement of $\Phi(y_i)$ with $\hat{\Phi}(x_i)$ facilitates an end-to-end learning. Because, in a non-differentiable model, $y_i$ is the input variable, while $x_i$ is regarded as the input of DNNs in differentiable models. 
\textbf{2)} We use $\mathop{\arg\max}$ to obtain the prediction, which is consistent with the nature of DNNs. Namely, during the testing phase, we directly obtain the prediction from the output of the DNN without the graphical losses. \textbf{3)} We use a differentiable term $-(\mathbf{Y}_i \mathbf{P}_i)^{\mathbf{T}} \log(\mathbf{Y}_j \mathbf{P}_j)$ to approximate the non-differential term ${\zeta (y_i, y_j)}{\mathbf{Y}_{i} \mathbf{Y}_{j}}$ in Eq. \eqref{eq:pairwise}.}

\subsection{Network Architecture and Loss functions}\label{sect:overall}
{The network architectures for training and testing are illustrated in Fig. \ref{fig:method}, where the black dotted lines denote training flow, and the solid black lines denote inference flow.}

Our weakly-supervised Re-ID model consists of three main modules, including \textbf{(a)} a feature embedding module built upon a ResNet-50 network followed by two fully connected layers, \textbf{(b)} a rough Re-ID module using a fully connected layer as the classifier, and \textbf{(c)} a refined Re-ID module that considers both the rough results and bag-level weak annotation to perform graphical modeling. %These modules are shown in Fig. \ref{fig:method}.
{It is noteworthy that we perform graphical modeling only in the training stage for two reasons. \textbf{First}, the graphical module is introduced to generate pseudo labels to supervise the model training, which requires a bag-level label as a constraint. However, there is no bag-level label in the testing stage. \textbf{Second}, due to the specificity of the Re-ID problem, the images in the inference stage are not organized in the form of a bag. For example, only a query image and a set of gallery images are provided in inference. As a result, there is no bag-level dependency among the testing images to exploit. Thus, performing graphical modeling may be infeasible in the inference stage. In the following, we will elaborate the three main modules.}

\textbf{Feature embedding module.} Many current best-performing Re-ID models use multi-scale features as feature embeddings \cite{sun2018beyond}, which guarantees a robust feature representation and thus boosts the performance. However, in this work, our focus is the mechanism of the weakly supervised Re-ID model alone, rather than other tricks. Therefore, we simply take the ResNet-50 \cite{he2016deep} as the backbone without any feature pyramid \cite{sun2018beyond}. Our feature embedding is similar to \cite{hermans2017defense}. Specifically, the last layer of the original ResNet-50 is discarded, and two new, fully connected layers are added. The first has $512$ units, followed by a batch normalization \cite{ioffe2015batch}, a Leaky ReLU \cite{nair2010rectified} and a dropout \cite{srivastava2014dropout}. This module is shown in Fig. \ref{fig:method} (a).

{\textbf{Rough Re-ID module.} We utilize a softmax classifier for rough Re-ID. Specifically, our model has a fully connected layer at the top of the feature embedding module, which has the same number of units to that of person ID (i.e., `class num' in Fig. \ref{fig:method}). A softmax cross-entropy loss is employed for training. The derived person categorization score (e.g., $\textbf{P}$ in Fig. \ref{fig:method}) is considered as the rough Re-ID estimation, indicating the possibility of a person ID being present in a bag. This module is shown in Fig. \ref{fig:method} (b).}

{\textbf{Refined Re-ID module.} Here, we aim to generate a pseudo image-level label $\hat{y}$ for each image by refining the previous estimation results. The refinement has the following inputs: \begin{enumerate}
\item{\emph{Rough Re-ID score.} As mentioned above, the rough Re-ID module provides a preliminary categorization.}
\item{\emph{Appearance.} 
Considering that rough Re-ID score is just a high-level abstraction of images, as compensation, we propose to integrate person appearance as low-level information for our refinement.}
\item{\emph{Bag constraint.} Intuitively, our bag constraint eliminates any possibility of assigning a person image with a person ID that is absent in the bag-level annotation; on the contrast, it encourages a person image to be assigned with a person ID that is present in the bag-level annotation. }
\end{enumerate}
}

{Accordingly, once the refined pseudo labels are generated, they are used to update the network weights as authentic ground truth:
\begin{small}\begin{equation}\label{eq:cls}
\mathcal{L}_{cls} = -\sum_{i=1}^{n}  (h(\hat{y}_{i}))^{\mathbf{T}} \log (\mathbf{P}_i).
\end{equation}\end{small}By combining Eq. \eqref{eq:cls} and Eq. \eqref{eq:crfloss}, we have the final loss function:\begin{small}\begin{equation}\label{eq:loss}
\mathcal{L} = w_{cls} \mathcal{L}_{cls} + w_{graph} \mathcal{L}_{graph},
\end{equation}\end{small}where $w_{cls}$ and $w_{graph}$ weights two loss components, respectively. In our experiments, we search the loss weight in a grid of \{1:1, 1:0.5, 1:0.1\} and find that 1:0.5 has good results on the CUHK03 dataset. 
}

\subsection{{Weakly-Supervised Triplet Loss}}\label{sect:detail}
{To further improve the performance of the Re-ID model, inspired by multiple granulariry network (MGN) \cite{Wang2018Wang_mm}, we propose a weakly-supervised triplet loss and derive our weakly supervised MGN (W-MGN) for weakly supervised Re-ID. MGN are learned with a triplet loss given strong annotations, while our weakly-supervised triplet loss in W-MGN is developed to address the strict dependency on annotations.
}

\vspace{3pt}
{Recall that a fully supervised triplet loss is defined as:
\begin{scriptsize}\begin{equation}\label{eq:ftri}
\begin{aligned}
 \mathcal{L}_{full{\_}triplet} = \displaystyle\sum_{k=1}^{tp}\bigg[\Delta &~+ \max_{\substack{j\ne k; j=1,\dotso, tp; y_k = y_j }}||z_k - z_j||_2^2\\
          & -\min_{\substack{j\neq k; j=1,\dotso, tp; y_k \neq y_j }}||z_k-z_j||_2^2\bigg],
\end{aligned} \end{equation}\end{scriptsize}where $t$ denotes the number of bags in a training batch, $||\cdot||_2^2$ denotes a $L_2$-norm, and $\Delta$ is a margin. The image-level labels $y_k$ and $y_j$ are given for the fully supervised triplet loss, but they are unavailable in the weakly supervised scenario.
}

\vspace{3pt}
{To address this problem, we use $\mathbf{Y}_k^T\mathbf{Y}_j > 0$ to approximate the constraint that $y_k = y_j$ in Eq. \eqref{eq:ftri}, which means that the $k$-th and $j$-th samples belong to the same class. As $\mathbf{Y}_k^T\mathbf{Y}_j > 0$ is a necessary but not sufficient condition of the constraint that $y_k = y_j$, we relax the $\max$ operation to a $\mathop{\text{median}}$ operation. Similarly, we use $\mathbf{Y}_k^T\mathbf{Y}_j = 0$ to approximate $y_k \neq y_j$. Accordingly, our weakly-supervised triplet loss is formulated as:
\begin{scriptsize}\begin{equation}\label{eq:wtri}
\begin{aligned}
 \mathcal{L}_{weak{\_}triplet} = \displaystyle\sum_{k=1}^{tp}\bigg[\Delta &~+ \mathop{\text{median}}_{\substack{j\ne k; j=1,\dotso, tp; \mathbf{Y}_k^T\mathbf{Y}_j > 0}}||z_k - z_j||_2^2\\
          & -\min_{\substack{j\neq k; j=1,\dotso, tp; \mathbf{Y}_k^T\mathbf{Y}_j =0}}||z_k-z_j||_2^2\bigg].
\end{aligned} \end{equation}\end{scriptsize}
Finally, our loss function is 
\begin{small}\begin{equation}\label{eq:loss}
\mathcal{L} = w_{cls} \mathcal{L}_{cls} + w_{graph} \mathcal{L}_{graph} + w_{triplet} \mathcal{L}_{weak\_triplet},
\end{equation}\end{small} where $w_{triplet}$ is a weight of our triplet loss.
}

\begin{figure*}[t]
\centering
\subfigure{
\centering
\includegraphics[width=0.9\linewidth]{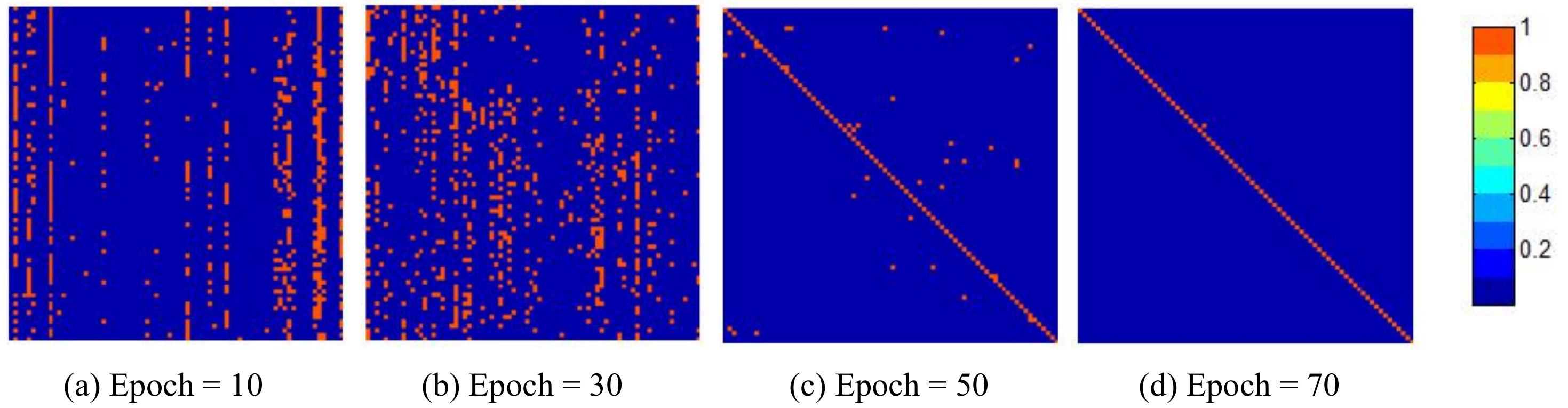}
}
\vspace{-11pt}
\caption{{\small{The effectiveness of our differentiable graphical learning module. Here we show the errors between the rough predictions and the weak annotations in the form of a confusion matrix containing $76\times76$ grids. Each grid indicates a bag of 10 categories, with a total sum of 760 categories, which is approximately equivalent to the person categories in the full training set (i.e., 767 categories).}}}\label{fig:refinement}
\vspace{-11pt}
\end{figure*}

\subsection{Computational Complexity}\label{sect:complexity}
{We discuss the computational cost of our weakly supervised Re-ID model. 
In the training phase, the extra time cost only relates to the generation of pseudo labels, which is a graphical learning module. Generally, graphical learning needs many iterations to search the optimal solution, and thus, the process is time-consuming. However, our approach formulates the differentiable graphical learning as a simple loss. This makes our graphical module very effective. In the experiment section, we will show that our training brings an additional time cost of only $0.002\times$. Particularly, in the testing phase, there is no extra time cost because the pseudo label generation component is disabled. In brief, this extra time cost of our method is negligible.}

\section{Relationship to Previous Works}\label{sect:relationship}
In the following, we compare our weakly supervised Re-ID with previous works on Re-ID with uncertain labels, including the unsupervised/semi-supervised Re-ID. In general, our weakly supervised Re-ID possesses not only cheap annotation but also high accuracy. The details are presented below.

\textbf{Unsupervised Re-ID.} To get rid of the prohibitively high cost of manual labeling, unsupervised learning Re-ID proposes to use either local saliency matching models or clustering models \cite{yu2017cross}. However, without the help of labeled data, it is difficult to model the dramatic variances across camera views in representation/metric learning. Therefore, it is difficult for these pipelines to obtain high accuracies \cite{song2018unsupervised,yu2019unsupervised,fan2018unsupervised,Li2019Unsupervised_tpami}. In contrast, our weakly supervised Re-ID problem has a better solution. Note that compared to unsupervised Re-ID, the annotation effort of our weakly supervised Re-ID is also very inexpensive.

\textbf{Semi-supervised Re-ID.} One-shot/one-example \cite{wu2019progressive,wu2018exploit} propose to reduce the annotation effort by annotating only one example for each person ID. The main differences between their methods and ours are two-fold. First, in one-shot Re-ID, at least one accurate label for each person category is in desire. While in our weakly supervised Re-ID, no accurate label is needed. Second, there is a bag-level label as a constraint to estimate the pseudo labels in our method, ensuring that our generated pseudo labels to be more reliable than those generated by one-shot Re-ID. 

We would also like to acknowledge the contribution of previous work \cite{meng2019weakly} that matches a target person image with a bag-level gallery video using multiple-instance multiple-label learning. However, similar to \cite{wu2018exploit}, at least one accurate label (of the target person) for each person category is still in a desire to form the probe set in \cite{meng2019weakly}. Hence, mathematically, \cite{meng2019weakly} still belongs to semi-supervised Re-ID but NOT weakly supervised Re-ID.

Section \ref{sect:cuhk03} and \ref{sect:market} will compare the accuracy of our weakly supervised Re-ID with previous works.

\begin{table*}[t]
\caption{\small{{Ablation studies of the proposed weakly supervised Re-ID method. \textbf{random:} Each bag contains random person IDs, which reflects the real-world state. \textbf{RK:} re-ranking, see \cite{zhong2017re}, one of the effective tricks frequently used in fully supervised Re-ID problems. $\mathbf{\star}$ \textbf{fully supervised:} when each bag contains only one person ID, the weakly supervised Re-ID problem degrades into a fully supervised Re-ID problem. $\mathbf{\ast}$\textbf{full training set:} the overall training set of CUHK03 contains 767 person IDs. \textbf{w/:} with. \textbf{w/o:} without.}}}\label{tab:ablation}
\centering
\begin{tabular}{ccccccccccc}
\multicolumn{5}{c}{(a) Impact of bag diversity on PRID2011} & \multicolumn{1}{c}{} & \multicolumn{5}{c}{(b) Impact of bag diversity on CUHK03} \\ \cline{1-5} \cline{7-11}
\multicolumn{2}{|c|}{categories / bag} & \multicolumn{1}{c|}{Rank-1} & \multicolumn{1}{c|}{Rank-5} & \multicolumn{1}{c|}{Rank-10} & \multicolumn{1}{c|}{} & \multicolumn{2}{c|}{categories / bag} & \multicolumn{1}{c|}{Rank-1} & \multicolumn{1}{c|}{Rank-5} & \multicolumn{1}{c|}{Rank-10} \\ \cline{1-5} \cline{7-11}
\multicolumn{2}{|c|}{1 ($\mathbf{\star}$ \textbf{fully supervised})} & \multicolumn{1}{c|}{71.8} & \multicolumn{1}{c|}{91.2} & \multicolumn{1}{c|}{95.9} & \multicolumn{1}{c|}{} & \multicolumn{2}{c|}{1 ($\mathbf{\star}$ \textbf{fully supervised})} & \multicolumn{1}{c|}{67.5} & \multicolumn{1}{c|}{88.2} & \multicolumn{1}{c|}{91.8} \\ \cline{1-5} \cline{7-11}
\multicolumn{2}{|c|}{2} & \multicolumn{1}{c|}{68.0} & \multicolumn{1}{c|}{87.5} & \multicolumn{1}{c|}{94.8} & \multicolumn{1}{c|}{} & \multicolumn{2}{c|}{2} & \multicolumn{1}{c|}{61.0} & \multicolumn{1}{c|}{82.0} & \multicolumn{1}{c|}{87.0} \\ \cline{1-5} \cline{7-11}
\multicolumn{2}{|c|}{3} & \multicolumn{1}{c|}{66.1} & \multicolumn{1}{c|}{86.4} & \multicolumn{1}{c|}{92.3} & \multicolumn{1}{c|}{} & \multicolumn{2}{c|}{3} & \multicolumn{1}{c|}{59.4} & \multicolumn{1}{c|}{80.7} & \multicolumn{1}{c|}{86.7} \\ \cline{1-5} \cline{7-11}
\multicolumn{2}{|c|}{10} & \multicolumn{1}{c|}{49.5} & \multicolumn{1}{c|}{73.9} & \multicolumn{1}{c|}{82.2} & \multicolumn{1}{c|}{} & \multicolumn{2}{c|}{10} & \multicolumn{1}{c|}{55.2} & \multicolumn{1}{c|}{79.3} & \multicolumn{1}{c|}{84.5} \\ \cline{1-5} \cline{7-11}
\multicolumn{2}{|c|}{\textbf{random (5 on average)}} & \multicolumn{1}{c|}{69.3} & \multicolumn{1}{c|}{89.0} & \multicolumn{1}{c|}{94.0} & \multicolumn{1}{c|}{} & \multicolumn{2}{c|}{\textbf{random (5 on average)}} & \multicolumn{1}{c|}{60.6} & \multicolumn{1}{c|}{81.6} & \multicolumn{1}{c|}{87.0} \\ \cline{1-5} \cline{7-11}
 &  &  &  &  &  &  &  &  &  &  \\ 
\multicolumn{5}{c}{(c) Fully supervised learning tricks on PRID2011} & \multicolumn{1}{c}{} & \multicolumn{5}{c}{(d) Fully supervised learning tricks on CUHK03} \\ \cline{1-5} \cline{7-11}
\multicolumn{2}{|c|}{method} & \multicolumn{1}{c|}{Rank-1} & \multicolumn{1}{c|}{Rank-5} & \multicolumn{1}{c|}{Rank-10} & \multicolumn{1}{c|}{} & \multicolumn{2}{c|}{method} & \multicolumn{1}{c|}{Rank-1} & \multicolumn{1}{c|}{Rank-5} & \multicolumn{1}{c|}{Rank-10} \\ \cline{1-5} \cline{7-11}
\multicolumn{2}{|c|}{fully supervised alone} & \multicolumn{1}{c|}{48.9} & \multicolumn{1}{c|}{79.6} & \multicolumn{1}{c|}{88.8} & \multicolumn{1}{c|}{} & \multicolumn{2}{c|}{fully supervised alone} & \multicolumn{1}{c|}{52.1} & \multicolumn{1}{c|}{77.9} & \multicolumn{1}{c|}{85.6} \\ \cline{1-5} \cline{7-11}
\multicolumn{2}{|c|}{weaky supervised alone} & \multicolumn{1}{c|}{39.9} & \multicolumn{1}{c|}{71.2} & \multicolumn{1}{c|}{83.3} & \multicolumn{1}{c|}{} & \multicolumn{2}{c|}{weaky supervised alone} & \multicolumn{1}{c|}{44.0} & \multicolumn{1}{c|}{70.6} & \multicolumn{1}{c|}{79.7} \\ \cline{1-5} \cline{7-11}
\multicolumn{2}{|c|}{fully supervised + \textbf{RK}} & \multicolumn{1}{c|}{71.8} & \multicolumn{1}{c|}{91.2} & \multicolumn{1}{c|}{95.9} & \multicolumn{1}{c|}{} & \multicolumn{2}{c|}{fully supervised + \textbf{RK}} & \multicolumn{1}{c|}{67.5} & \multicolumn{1}{c|}{88.2} & \multicolumn{1}{c|}{91.8} \\ \cline{1-5} \cline{7-11}
\multicolumn{2}{|c|}{weakly supervised + \textbf{RK}} & \multicolumn{1}{c|}{68.0} & \multicolumn{1}{c|}{87.5} & \multicolumn{1}{c|}{94.8} & \multicolumn{1}{c|}{} & \multicolumn{2}{c|}{weakly supervised + \textbf{RK}} & \multicolumn{1}{c|}{61.0} & \multicolumn{1}{c|}{82.0} & \multicolumn{1}{c|}{87.0} \\ \cline{1-5} \cline{7-11}
 &  &  &  &  &  &  &  &  &  &  \\ 
 \multicolumn{5}{c}{(e) Scalability of our method on CUHK03} & \multicolumn{1}{c}{} & \multicolumn{5}{c}{(f) Effectiveness of the graphical learning module on CUHK03} \\ \cline{1-5} \cline{7-11}
\multicolumn{2}{|c|}{categories} & \multicolumn{1}{c|}{Rank-1} & \multicolumn{1}{c|}{Rank-5} & \multicolumn{1}{c|}{Rank-10} & \multicolumn{1}{c|}{} & \multicolumn{2}{c|}{method} & \multicolumn{1}{c|}{Rank-1} & \multicolumn{1}{c|}{Rank-5} & \multicolumn{1}{c|}{Rank-10} \\ \cline{1-5} \cline{7-11}
\multicolumn{2}{|c|}{67} & \multicolumn{1}{c|}{16.3} & \multicolumn{1}{c|}{34.7} & \multicolumn{1}{c|}{44.9} & \multicolumn{1}{c|}{} & \multicolumn{2}{c|}{w/o graphical model} & \multicolumn{1}{c|}{56.4} & \multicolumn{1}{c|}{80.0} & \multicolumn{1}{c|}{85.1} \\ \cline{1-5} \cline{7-11}
\multicolumn{2}{|c|}{367} & \multicolumn{1}{c|}{43.6} & \multicolumn{1}{c|}{67.0} & \multicolumn{1}{c|}{75.5} & \multicolumn{1}{c|}{} & \multicolumn{2}{c|}{w/o pairwise term} & \multicolumn{1}{c|}{59.2} & \multicolumn{1}{c|}{80.9} & \multicolumn{1}{c|}{86.7} \\ \cline{1-5} \cline{7-11}
\multicolumn{2}{|c|}{767 ($\mathbf{\ast}$\textbf{full training set})} & \multicolumn{1}{c|}{61.0} & \multicolumn{1}{c|}{82.0} & \multicolumn{1}{c|}{87.0} & \multicolumn{1}{c|}{} & \multicolumn{2}{c|}{w/ graphical model} & \multicolumn{1}{c|}{61.0} & \multicolumn{1}{c|}{82.0} & \multicolumn{1}{c|}{87.0} \\ \cline{1-5} \cline{7-11}
\end{tabular}
\vspace{-11pt}
\end{table*}

\section{Experiments}\label{sect:exp}
{In this section, we conducts extensive experiments to evaluate our weakly supervised Re-ID approach. Section \ref{sec:exp_setting} presents the experimental settings. Section \ref{sect:ablation_study} provides a comprehensive ablation study. Section \ref{sect:sota} presents comparisons of our approach with state-of-the-art methods and also provides the discussion on the computational cost.}

\subsection{Experimental settings}\label{sec:exp_setting}
\subsubsection{Datasets}
In addition to the proposed SYSU-$30k$ dataset, another four simulated datasets are introduced to evaluate the effectiveness of our method by adjusting the existing datasets. Specifically, we replace the strong annotations on the training set of the PRID2011 \cite{hirzer2011person}, CUHK03 \cite{li2014deepreid}, Market-1501 \cite{zheng2015scalable}, and MSMT17 \cite{wei2018person} with weak annotations while their test sets are kept unchanged, as the definition states that during testing, there is no difference between the fully and weakly supervised Re-ID (see Fig. \ref{fig:weakly} (c)). For a fair comparison (e.g., using the same images for both the fully and weakly supervised Re-ID), we generate the weak annotations from the strong annotations.
This includes two steps. {\textbf{First}}, each bag is simulated by randomly selecting several images and packaging them. {\textbf{Second}}, the weak labels are obtained by summarizing the strong annotations, e.g., four image-level labels \{Alice, Bob, Alice, Carol\} are summarized as a bag-level label \{Alice, Bob, Carol\}.
We denote $n$ ID/bag when a bag contains $n$ person IDs.
Note that unless otherwise stated, our weakly supervised learning setting is two IDs/bag.

\emph{PRID2011.} Originally, PRID2011 dataset contains 200 person IDs appearing in at least two camera views and is further randomly divided into training/test sets following the general settings \cite{wang2017p2snet}, i.e., both having 100 IDs.

{\emph{CUHK03.} CUHK03 is a large-scale Re-ID, which contains 14,096 images of 1,467 IDs collected from 5 different pairs of camera views \cite{li2014deepreid}. Each ID is observed by two disjointed camera views. We follow the \textbf{new} standard protocol \cite{zhong2017re} of CUHK03, i.e., a training set including 767 IDs is obtained without overlap.}

\emph{Market-1501.} Market-1501 is another widely-used large-scale Re-ID benchmark, which contains 32,668 images of 1,501 IDs captured from 6 different cameras. The dataset is split into two parts: 12,936 images with 751 IDs for training and 19,732 images with 750 IDs for testing. In testing, 3,368 hand-drawn images with 750 IDs are used as probe set to identify the true IDs on the testing set.

{\emph{MSMT17.} MSMT17 is the current largest publicly available Re-ID dataset and there are 126,441 images in MSMT17 in total captured by 15 cameras. It has 4$k$ person IDs, namely, 7.5 times smaller than our SYSU-30$k$. We follow the standard protocol to split the training and testing set \cite{wei2018person}.}
{\subsubsection{Implementation details}
The parameters of the ResNet-50 backbone are initialized using ImageNet pre-training. Other parameters are initialized by sampling from a normal distribution. For SGD, we use a minibatch of 90 images and an initial learning rate of 0.01 (0.1 for the fully connected layer), multiplying the learning rate by 0.1 after a fixed number of iterations. We use the momentum of 0.9 and a weight decay of 0.0005. Training on SYSU-{$30k$} takes approximately ten days on a single GPU (i.e., NVIDIA TITAN X).}

\subsection{Ablation Study} \label{sect:ablation_study}
We first present ablation studies to reveal the benefits of each main component of our method.

\subsubsection{Effectiveness of the graphical learning module} As aforementioned, the graphical learning module plays the role of refining the ID prediction by correcting the errors between the rough Re-ID predictions and the weak annotations, which forms the basis of generating pseudo-image-level labels. We visualize the errors between the rough predictions and the weak annotations in Fig. \ref{fig:refinement} during training. This experiment is conducted on CUHK03 using the setting of 10 IDs/bag.

{Fig. \ref{fig:refinement} shows the errors between the rough predictions and the weak annotations in the form of a confusion matrix containing $76\times76$ grids. Each grid indicates a bag of 10 IDs, totally summing up to 760 IDs, which approximates the number of person IDs in the full training set. We have two appealing observations from Fig. \ref{fig:refinement}. \textbf{First}, there is a significant gap between the rough predictions and the weak annotations (see \ref{fig:refinement} (a) or (b)), indicating that the rough Re-ID results are still not competent for generating pseudo labels. Therefore, refining the ID prediction is necessary with our graphical learning module. \textbf{Second}, the gap between the rough predictions and the weak annotations becomes smaller as the training iteration increases (from 10 epochs in \ref{fig:refinement}(a) to 70 epochs in \ref{fig:refinement} (d)). When the training model converges, the gap between the ground truth becomes significantly small, which indicates that the problem is well addressed by our graphical learning module.}

{To further demonstrate the effectiveness of the differential graphical models, we conduct two more empirical studies on CUHK03 and Market-1501 with and without the graphical model, respectively. The graphical module generates pseudo labels to further supervise the learning. Once it is removed, there will be no pseudo labels provided. To cope with this problem, we use the preliminary image-level labels in Eq. \eqref{eq:prob} as substitutes for pseudo labels, which is also a widely used strategy in existing weakly supervised learning. Table \ref{tab:ablation} (f) shows that without the graphical model, there is a significant performance drop on CUHK03, i.e., from 61.0\% to 56.4\%. Similarly, the performance of W-MGN drops significantly from 95.5\% to 88.4\% when the graphical model is removed, as shown in Table \ref{tab:cmp_sota} (a). 
These comparisons clearly demonstrate the effectiveness of our graphical model.}

\subsubsection{Effectiveness of the pairwise term} {As claimed above, the pairwise term is necessary for label smoothness in order to train better models. To validate the necessity of the pairwise term, we conduct two more experiments on CUHK03 and Market-1501 with and without the pairwise term, respectively. Without pairwise terms, a graphical model reduces to isolated nodes. Experimental results are provided in Table \ref{tab:ablation} (f) and Table \ref{tab:cmp_sota} (a). For example, on Market-1501 test set, the accuracy drops significantly from 95.5\% to 94.0\% without the pairwise term.
}

\begin{figure}
\centering
\subfigure[\small{PRID2011}]{
  \centering
  \includegraphics[width=0.46\linewidth]{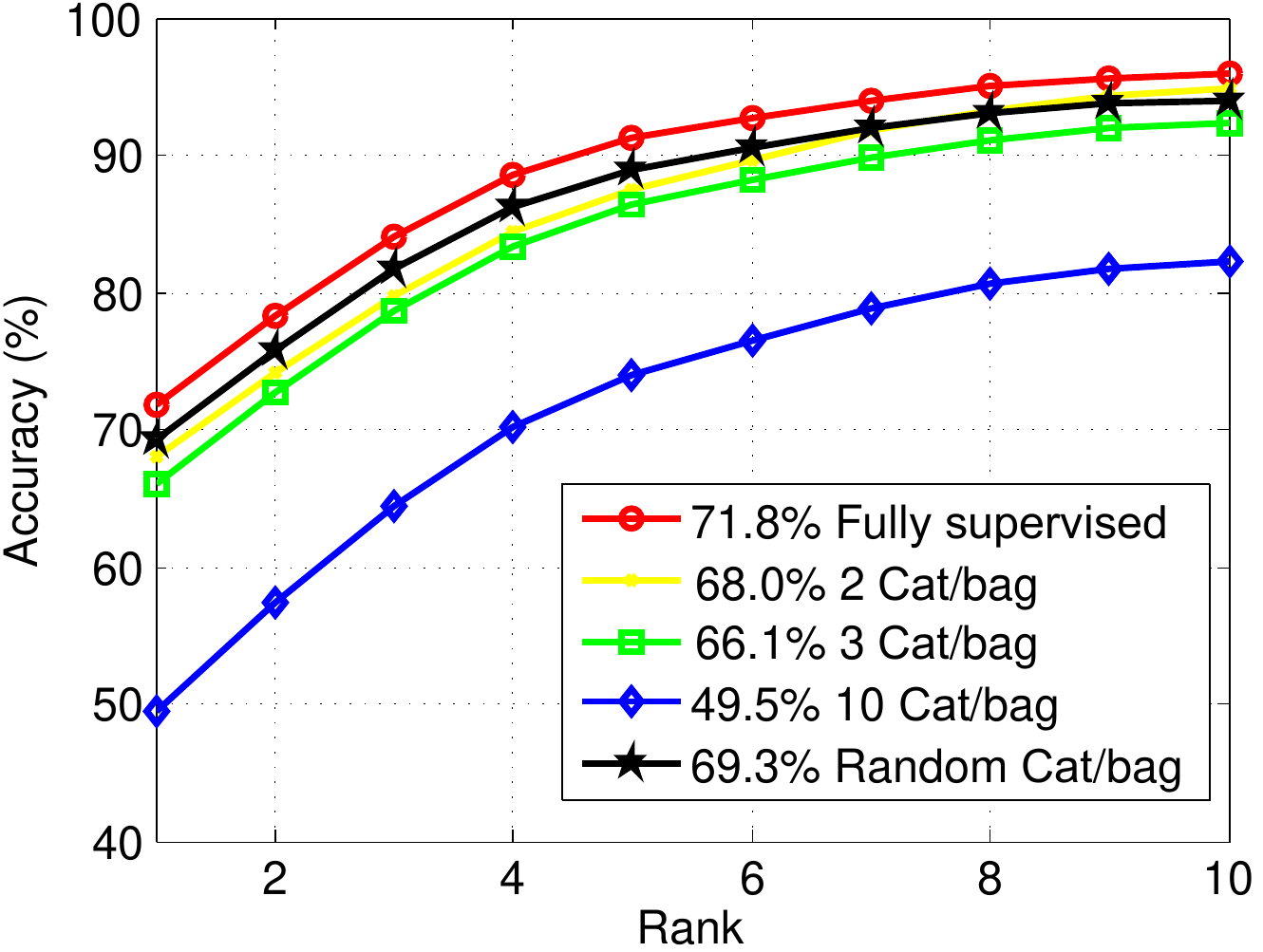}
  }
\subfigure[\small{CUHK03}]{
  \centering
  \includegraphics[width=0.46\linewidth]{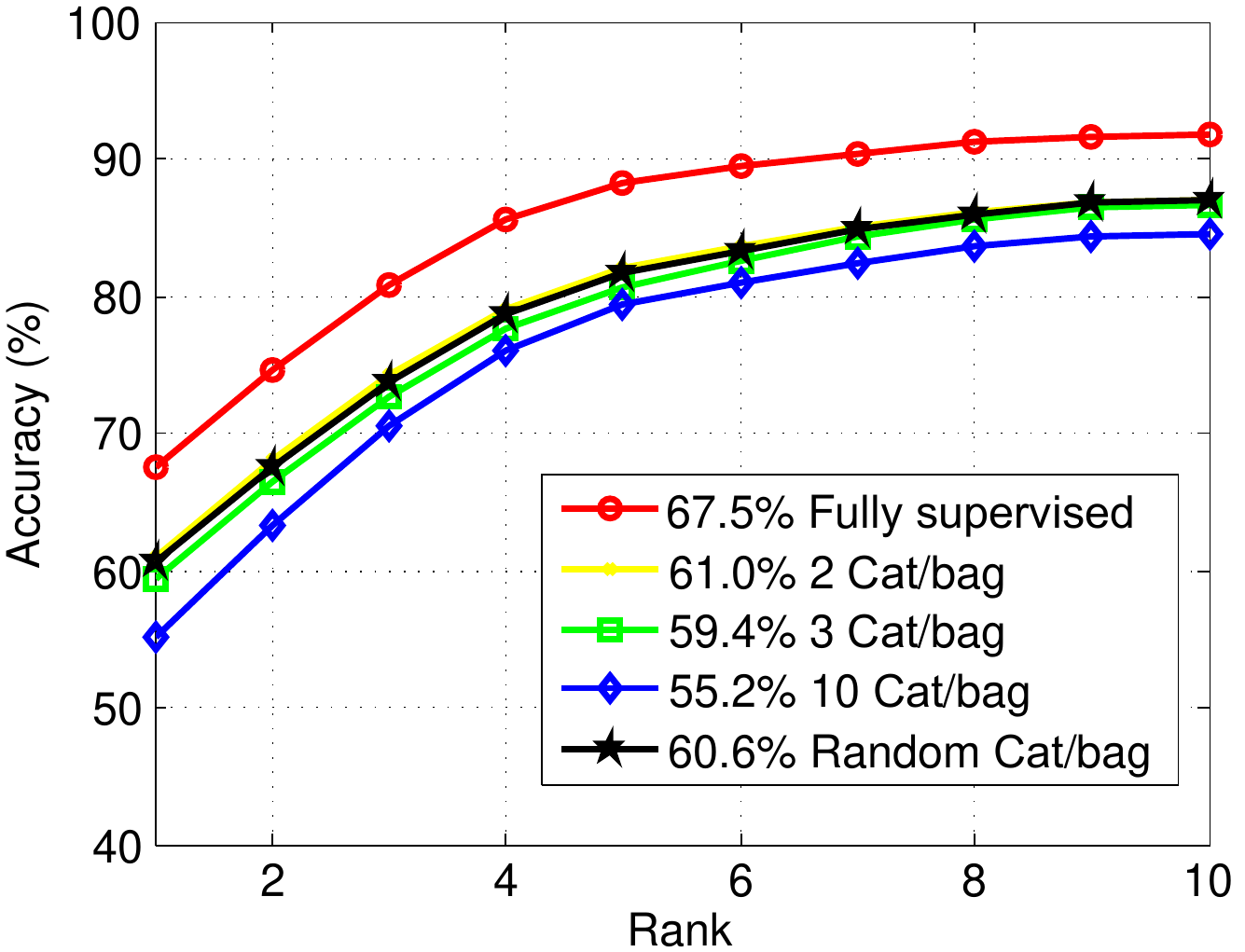}
  }
  \vspace{-11pt}
\caption{\small{Analysis on different bag diversities. \textbf{Cat/bag:} the number of person IDs in each bag. \textbf{Random Cat/bag:} each bag contains random number of person IDs, which reflects the real-world state. \textbf{Fully supervised:} each bag contains only one person ID. In this case, the weakly supervised problem degrades into a fully supervised one.}}\label{fig:diversity}
\vspace{-11pt}
\end{figure}

\subsubsection{Effectiveness of the weakly supervised triplet loss} {
To validate the effectiveness, we conduct ablation studies on our weakly-supervised triplet loss on five benchmarks (i.e., Market-1501, CUHK03, PRID2011, MSMT17, and SYSU-30$k$), respectively.
Table \ref{tab:cmp_sota} (a)-(e) show that W-MGN with the weakly-supervised triplet loss outperforms that without the weakly-supervised triplet loss ((i.e., ``W-MGN w/o W-Tr'')) by a large margin on all the five datasets, which verifies its effectiveness. For example, without the weakly-supervised triplet loss, W-MGN suffers a performance degradation from 95.5\% to 92.9\% on Market-1501.}

\subsubsection{Effectiveness of stronger baselines} {To see how stronger baselines affect the performance of the weakly supervised Re-ID, we compare more strong baselines like Local CNN \cite{Yang2018Local_mm} and MGN \cite{Wang2018Wang_mm} with our ResNet-50 baseline on the five datasets, i.e., Market-1501, CUHK03, PRID2011, MSMT17, and SYSU-30$k$. Table \ref{tab:cmp_sota} (a)-(e) show that strong baselines for fully supervised Re-ID contribute to the overall performance of the weakly supervised method on all the five datasets. For example, the combination of our weakly supervised Re-ID and MGN (i.e., W-MGN) outperforms the combination of the weakly supervised Re-ID and ResNet-50 (i.e., W-Baseline) by 6.9\% (95.5\% \emph{vs.} 88.6\%).}

\subsubsection{Scalability of our approach}

We have shown that a Re-ID model can be learned with weakly labeled data. Next, we investigate whether increasing the amount of weakly labeled data will improve the performance of weakly supervised learning. The entire CUHK03 training set is randomly partitioned into three subsets containing 67, 300, and 300 person IDs, respectively. We evaluate the scalability of our approach by gradually adding one subset in training. The rank-1 accuracy is reported in Table \ref{tab:ablation} (e). For example, the first model is trained with the first 67 person IDs, and the number of person IDs is increased to 367 IDs in the second model. The third model is trained with the full CUHK03 training set (i.e., 767 IDs). Table \ref{tab:ablation} (e) shows that the accuracies increase when we increase the scale of training data in CUHK03. For instance, our approach trained with full training data achieves the best performance and outperforms the other two models by 44.7\% and 17.4\%, respectively.

\subsubsection{Impact of bag diversity} {Intuitively, if a bag contains more person IDs, it is more challenging to learn a weakly supervised Re-ID model because of the increase in uncertainty. Next, we investigate the performance with respect to such bag internal diversity. We conduct experiments on PRID2011 and CUHK03. In Table \ref{tab:ablation} (a)-(b) and Fig. \ref{fig:diversity} (a)-(b), we compare five options, i.e., each bag containing 1, 2, 3, 10, or a random number of person IDs, respectively. In particular, when each bag has only one person ID, the weakly supervised Re-ID problem degrades into a fully supervised one.}

We have three major observations from Table \ref{tab:ablation} (a)-(b) and Fig. \ref{fig:diversity} (a)-(b). \textbf{First}, the models trained with weakly labeled samples achieves comparable accuracies to the models trained with strongly labeled data (e.g., 68.0\% \emph{vs.} 71.8\% in Table \ref{tab:ablation} (a)). This result is quite important because a weak annotation costs much less money and time than a strong annotation.

\textbf{Second}, the accuracy of the weakly supervised methods gradually decreases as the number of IDs in each bag increases. In particular, the rank-1 accuracy of our approach drops by 18.5\% when increasing the number of IDs per bag from 2 to 10 in Table \ref{tab:ablation} (a). We argue that the increase in uncertainty causes this optimization difficulty. When the IDs per bag increases, the uncertainty in the label assignment also increases, making the problem more challenging. 

{\textbf{Third}, it is noteworthy that the random version has appealing performance (69.3\% \emph{vs} 71.8\% compared with the baseline), as shown in the last line of Table \ref{tab:ablation} (a). Specifically, the random version refers to each bag containing a random number of person IDs (5 IDs on average), which reflects the real-world states. The high performance suggests that solving a weakly supervised Re-ID problem is feasible and appealing in reality.}

\begin{figure}
\centering
\subfigure[\small{Standard PRID2011 test set.}]{
  \centering
  \includegraphics[width=0.46\linewidth]{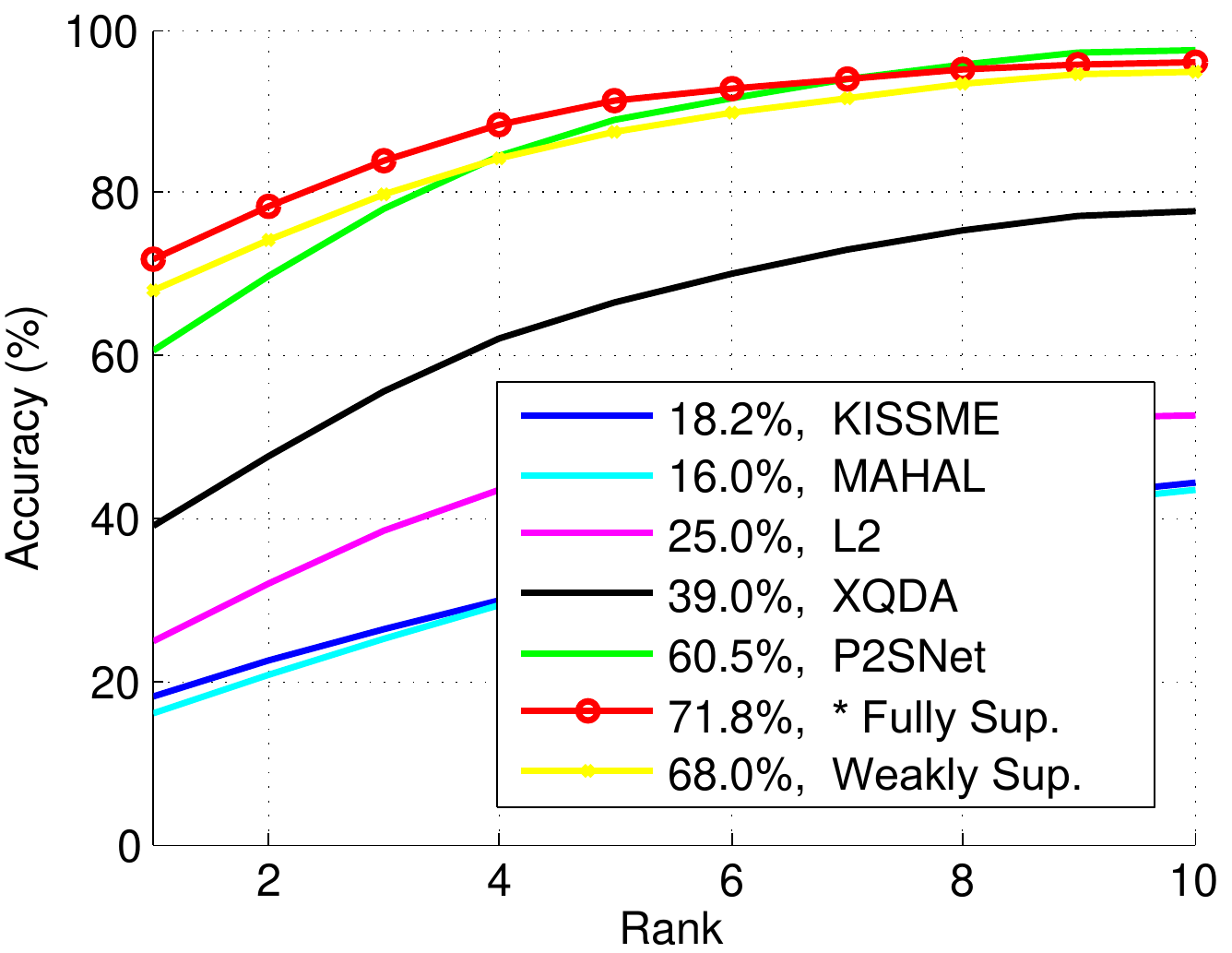}
  }
\subfigure[\small{SYSU-$30k$}]{
  \centering
  \includegraphics[width=0.46\linewidth]{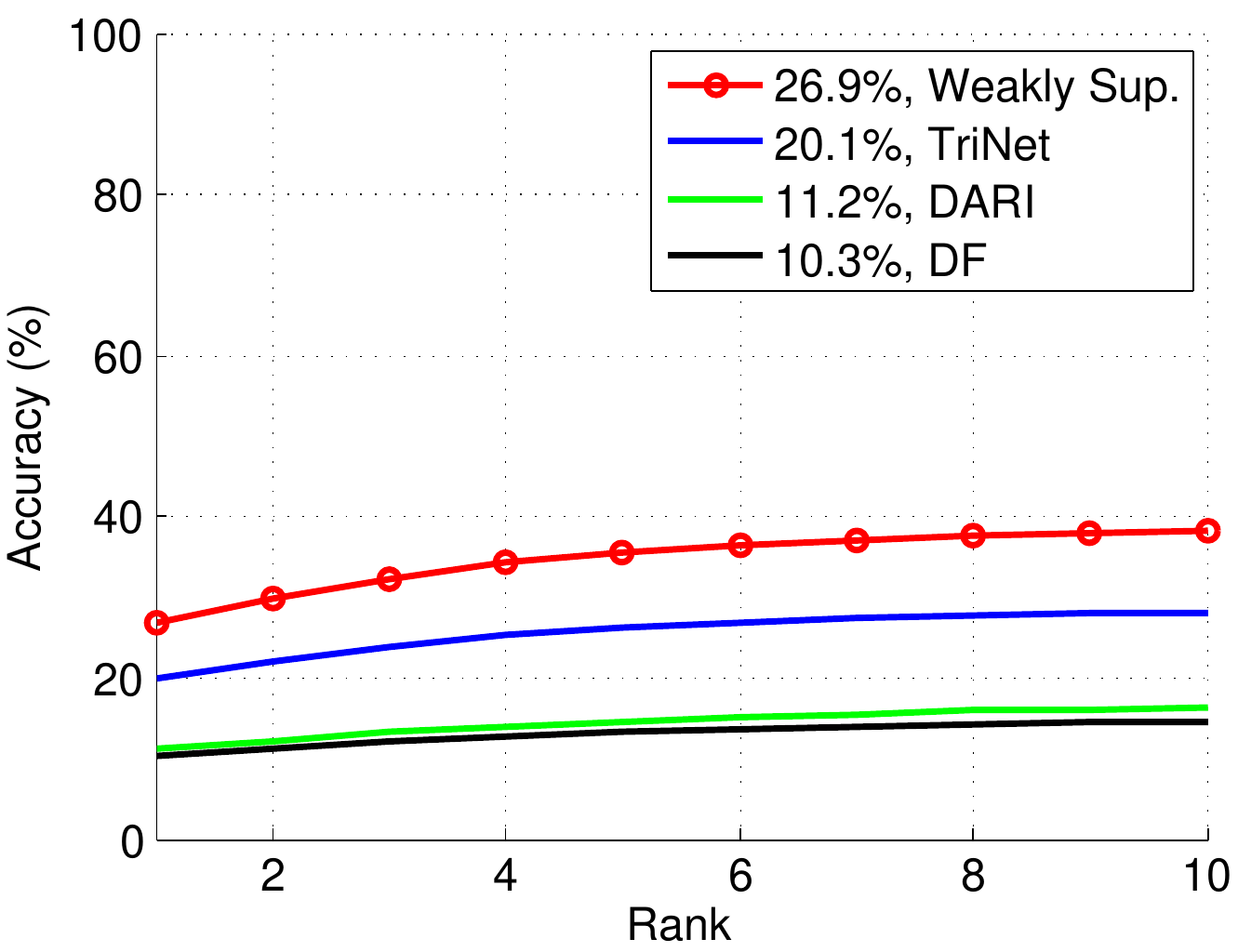}
  }
  \vspace{-11pt}
\caption{\small{Comparison with state-of-the-art methods. \textbf{Weakly sup.:} the proposed weakly supervised Re-ID approach.
\textbf{* Fully Sup.:} each bag contains only one person ID. In this case, the weakly supervised problem degrades into a fully supervised one.}}\label{fig:cmp_sota}
\vspace{-11pt}
\end{figure}

\begin{table*}
\vspace{-11pt}
\caption{\small{{Comparison with state-of-the-art methods. \textbf{W-Baseline:} our weakly supervised Re-ID method.
\textbf{Baseline:} each bag contains only one person ID. In this case, the weakly supervised problem degrades into a fully supervised one. We thus consider the latter as the baseline of our weakly supervised Re-ID. \textbf{w/o Tri:} without triplet loss. \textbf{x(y):} y is the number reported by the original paper; x is the result of our reproduction. \textbf{RK:} re-ranking. \textbf{\ddag:} pretrained on CUHK03. \textbf{w/:} with. \textbf{w/o:} without. \textbf{W-:} weakly supervised version of a method.}}}
\centering
\scriptsize
\begin{tabular}{ccccccccccc}
\multicolumn{3}{c}{(a) Market-1501 test set.}                                                                                                           &                       & \multicolumn{3}{c}{(b) CUHK03 test set.}                                                                                                              &                       & \multicolumn{3}{c}{(d) MSMT17 test set.}                                                                                                                      \\ \cline{1-3} \cline{5-7} \cline{9-11} 
\multicolumn{1}{|c|}{Supervision}                    & \multicolumn{1}{c|}{Method}                                        & \multicolumn{1}{c|}{Rank-1} & \multicolumn{1}{c|}{} & \multicolumn{1}{c|}{Supervision}                   & \multicolumn{1}{c|}{Method}                                        & \multicolumn{1}{c|}{Rank-1} & \multicolumn{1}{c|}{} & \multicolumn{1}{c|}{Supervision}                   & \multicolumn{1}{c|}{Method}                                                & \multicolumn{1}{c|}{Rank-1} \\ \cline{1-3} \cline{5-7} \cline{9-11} 
\multicolumn{1}{|c|}{\multirow{16}{*}{Fully}}        & \multicolumn{1}{c|}{MSCAN\cite{li2017learning}}                    & \multicolumn{1}{c|}{80.3}   & \multicolumn{1}{c|}{} & \multicolumn{1}{c|}{\multirow{15}{*}{Fully}}       & \multicolumn{1}{c|}{BOW+XQDA \cite{zheng2015scalable}}             & \multicolumn{1}{c|}{6.4}    & \multicolumn{1}{c|}{} & \multicolumn{1}{c|}{\multirow{17}{*}{Fully}}       & \multicolumn{1}{c|}{PDC\cite{Su2017Pose_iccv}}                             & \multicolumn{1}{c|}{58}     \\ \cline{2-3} \cline{6-7} \cline{10-11} 
\multicolumn{1}{|c|}{}                               & \multicolumn{1}{c|}{DF\cite{zhao2017deeply}}                       & \multicolumn{1}{c|}{81.0}   & \multicolumn{1}{c|}{} & \multicolumn{1}{c|}{}                              & \multicolumn{1}{c|}{PUL \cite{fan2018unsupervised}}                & \multicolumn{1}{c|}{9.1}    & \multicolumn{1}{c|}{} & \multicolumn{1}{c|}{}                              & \multicolumn{1}{c|}{GLAD\cite{Wei2017GLAD_acmmm}}                          & \multicolumn{1}{c|}{61.4}   \\ \cline{2-3} \cline{6-7} \cline{10-11} 
\multicolumn{1}{|c|}{}                               & \multicolumn{1}{c|}{SSM\cite{bai2017scalable}}                     & \multicolumn{1}{c|}{82.2}   & \multicolumn{1}{c|}{} & \multicolumn{1}{c|}{}                              & \multicolumn{1}{c|}{LOMO+XQDA \cite{liao2015person}}               & \multicolumn{1}{c|}{12.8}   & \multicolumn{1}{c|}{} & \multicolumn{1}{c|}{}                              & \multicolumn{1}{c|}{ABD-Net\cite{Chen2019ABD_iccv}}                        & \multicolumn{1}{c|}{82.3}   \\ \cline{2-3} \cline{6-7} \cline{10-11} 
\multicolumn{1}{|c|}{}                               & \multicolumn{1}{c|}{SVDNet\cite{sun2017svdnet}}                    & \multicolumn{1}{c|}{82.3}   & \multicolumn{1}{c|}{} & \multicolumn{1}{c|}{}                              & \multicolumn{1}{c|}{IDE(R) \cite{zheng2016person}}                 & \multicolumn{1}{c|}{21.3}   & \multicolumn{1}{c|}{} & \multicolumn{1}{c|}{}                              & \multicolumn{1}{c|}{GoogleNet\cite{Szegedy2015Going_cvpr}}                 & \multicolumn{1}{c|}{47.6}   \\ \cline{2-3} \cline{6-7} \cline{10-11} 
\multicolumn{1}{|c|}{}                               & \multicolumn{1}{c|}{GAN\cite{zheng2017unlabeled}}                  & \multicolumn{1}{c|}{84.0}   & \multicolumn{1}{c|}{} & \multicolumn{1}{c|}{}                              & \multicolumn{1}{c|}{IDE+DaF \cite{yu2017divide}}                   & \multicolumn{1}{c|}{26.4}   & \multicolumn{1}{c|}{} & \multicolumn{1}{c|}{}                              & \multicolumn{1}{c|}{IANet\cite{Hou2019Interaction_cvpr}}                   & \multicolumn{1}{c|}{75.5}   \\ \cline{2-3} \cline{6-7} \cline{10-11} 
\multicolumn{1}{|c|}{}                               & \multicolumn{1}{c|}{PDF\cite{Su2017Pose_iccv}}                          & \multicolumn{1}{c|}{84.1}   & \multicolumn{1}{c|}{} & \multicolumn{1}{c|}{}                              & \multicolumn{1}{c|}{IDE+XQ.+\textbf{RK}\cite{zhong2017re}}         & \multicolumn{1}{c|}{34.7}   & \multicolumn{1}{c|}{} & \multicolumn{1}{c|}{}                              & \multicolumn{1}{c|}{SFT\cite{Luo2019Spectral_iccv}}                        & \multicolumn{1}{c|}{73.6}   \\ \cline{2-3} \cline{6-7} \cline{10-11} 
\multicolumn{1}{|c|}{}                               & \multicolumn{1}{c|}{TriNet\cite{hermans2017defense}}               & \multicolumn{1}{c|}{84.9}   & \multicolumn{1}{c|}{} & \multicolumn{1}{c|}{}                              & \multicolumn{1}{c|}{PAN}                                           & \multicolumn{1}{c|}{36.3}   & \multicolumn{1}{c|}{} & \multicolumn{1}{c|}{}                              & \multicolumn{1}{c|}{Auto-ReID\cite{Quan2019Auto_iccv}}                     & \multicolumn{1}{c|}{78.2}   \\ \cline{2-3} \cline{6-7} \cline{10-11} 
\multicolumn{1}{|c|}{}                               & \multicolumn{1}{c|}{TriNet+Era.+\textbf{RK}\cite{zhong2017random}} & \multicolumn{1}{c|}{85.5}   & \multicolumn{1}{c|}{} & \multicolumn{1}{c|}{}                              & \multicolumn{1}{c|}{DPFL \cite{chen2017person}}                    & \multicolumn{1}{c|}{40.7}   & \multicolumn{1}{c|}{} & \multicolumn{1}{c|}{}                              & \multicolumn{1}{c|}{MVP\cite{Chen2019MVP_iccv}}                            & \multicolumn{1}{c|}{71.3}   \\ \cline{2-3} \cline{6-7} \cline{10-11} 
\multicolumn{1}{|c|}{}                               & \multicolumn{1}{c|}{PCB\cite{sun2018beyond}}                       & \multicolumn{1}{c|}{93.4}   & \multicolumn{1}{c|}{} & \multicolumn{1}{c|}{}                              & \multicolumn{1}{c|}{TreeConv\cite{Wang2020Grammatically_kdd}}                   & \multicolumn{1}{c|}{71.4}   & \multicolumn{1}{c|}{} & \multicolumn{1}{c|}{}                              & \multicolumn{1}{c|}{Verif-Identif\cite{Zheng2018Discriminatively_tomccap}} & \multicolumn{1}{c|}{60.5}   \\ \cline{2-3} \cline{6-7} \cline{10-11} 
\multicolumn{1}{|c|}{}                               & \multicolumn{1}{c|}{VPM\cite{sun2019perceive}}                     & \multicolumn{1}{c|}{93.0}   & \multicolumn{1}{c|}{} & \multicolumn{1}{c|}{}                              & \multicolumn{1}{c|}{TriNet+Era. \cite{zhong2017random}}            & \multicolumn{1}{c|}{55.5}   & \multicolumn{1}{c|}{} & \multicolumn{1}{c|}{}                              & \multicolumn{1}{c|}{PCB\cite{sun2018beyond}}                               & \multicolumn{1}{c|}{68.2}   \\ \cline{2-3} \cline{6-7} \cline{10-11} 
\multicolumn{1}{|c|}{}                               & \multicolumn{1}{c|}{JDGL\cite{zheng2019joint}}                     & \multicolumn{1}{c|}{94.8}   & \multicolumn{1}{c|}{} & \multicolumn{1}{c|}{}                              & \multicolumn{1}{c|}{ACNet\cite{wang2019adaptively}} & \multicolumn{1}{c|}{64.8}   & \multicolumn{1}{c|}{} & \multicolumn{1}{c|}{}                              & \multicolumn{1}{c|}{JDGL\cite{zheng2019joint}}                             & \multicolumn{1}{c|}{77.2}   \\ \cline{2-3} \cline{6-7} \cline{10-11} 
\multicolumn{1}{|c|}{}                               & \multicolumn{1}{c|}{AANet\cite{Tay_2019_CVPR}}                     & \multicolumn{1}{c|}{92.4}   & \multicolumn{1}{c|}{} & \multicolumn{1}{c|}{}                              & \multicolumn{1}{c|}{\textbf{Baseline}}                    & \multicolumn{1}{c|}{67.5}   & \multicolumn{1}{c|}{} & \multicolumn{1}{c|}{}                              & \multicolumn{1}{c|}{ShuffleNet\cite{Zhang2018ShuffleNet_cvpr}}             & \multicolumn{1}{c|}{41.5}   \\ \cline{2-3} \cline{6-7} \cline{10-11} 
\multicolumn{1}{|c|}{}                               & \multicolumn{1}{c|}{Local CNN \cite{Yang2018Local_mm}}                               & \multicolumn{1}{c|}{ 95.9(97.0)}       & \multicolumn{1}{c|}{} & \multicolumn{1}{c|}{}                              & \multicolumn{1}{c|}{Local CNN \cite{Yang2018Local_mm}}                               & \multicolumn{1}{c|}{69.6}       & \multicolumn{1}{c|}{} & \multicolumn{1}{c|}{}                              & \multicolumn{1}{c|}{Local CNN \cite{Yang2018Local_mm}}                                       & \multicolumn{1}{c|}{82.9}       \\ \cline{2-3} \cline{6-7} \cline{10-11} 
\multicolumn{1}{|c|}{}                               & \multicolumn{1}{c|}{MGN \cite{Wang2018Wang_mm} }                                           & \multicolumn{1}{c|}{95.8(96.6)}       & \multicolumn{1}{c|}{} & \multicolumn{1}{c|}{}                              & \multicolumn{1}{c|}{MGN \cite{Wang2018Wang_mm} }                                           & \multicolumn{1}{c|}{70.4(68.0)}       & \multicolumn{1}{c|}{} & \multicolumn{1}{c|}{}                              & \multicolumn{1}{c|}{MGN \cite{Wang2018Wang_mm} }                                                   & \multicolumn{1}{c|}{83.1}       \\ \cline{2-3} \cline{6-7} \cline{10-11} 
\multicolumn{1}{|c|}{}                               & \multicolumn{1}{c|}{MGN w/o Tri}                          & \multicolumn{1}{c|}{93.4}       & \multicolumn{1}{c|}{} & \multicolumn{1}{c|}{}                              & \multicolumn{1}{c|}{MGN w/o Tri}                          & \multicolumn{1}{c|}{67.6}       & \multicolumn{1}{c|}{} & \multicolumn{1}{c|}{}                              & \multicolumn{1}{c|}{MGN w/o Tri }                                  & \multicolumn{1}{c|}{81.0}       \\ \cline{2-3} \cline{5-7} \cline{10-11} 
\multicolumn{1}{|c|}{}                               & \multicolumn{1}{c|}{\textbf{Baseline}}                     & \multicolumn{1}{c|}{94.2}   & \multicolumn{1}{c|}{} & \multicolumn{1}{c|}{\multirow{3}{*}{Unsupervised}} & \multicolumn{1}{c|}{CAMEL\cite{yu2017cross}}                       & \multicolumn{1}{c|}{31.9}   & \multicolumn{1}{c|}{} & \multicolumn{1}{c|}{}                              & \multicolumn{1}{c|}{MobileNetV2\cite{Sandler2018MobileNetV2_cvpr}}         & \multicolumn{1}{c|}{50.9}   \\ \cline{1-3} \cline{6-7} \cline{10-11} 
\multicolumn{1}{|c|}{\multirow{11}{*}{Unsupervised}} & \multicolumn{1}{c|}{CAMEL\cite{yu2017cross}}                       & \multicolumn{1}{c|}{54.5}   & \multicolumn{1}{c|}{} & \multicolumn{1}{c|}{}                              & \multicolumn{1}{c|}{PatchNet\cite{Yang_2019_CVPR}}                 & \multicolumn{1}{c|}{45.4}   & \multicolumn{1}{c|}{} & \multicolumn{1}{c|}{}                              & \multicolumn{1}{c|}{OSNet\cite{Zhou2019Omni_iccv}}                         & \multicolumn{1}{c|}{78.7}   \\ \cline{2-3} \cline{6-7} \cline{9-11} 
\multicolumn{1}{|c|}{}                               & \multicolumn{1}{c|}{TAUDL\cite{li2018unsupervised}}                & \multicolumn{1}{c|}{63.7}   & \multicolumn{1}{c|}{} & \multicolumn{1}{c|}{}                              & \multicolumn{1}{c|}{PAUL\cite{Yang_2019_CVPR}}                     & \multicolumn{1}{c|}{52.3}   & \multicolumn{1}{c|}{} & \multicolumn{1}{c|}{\multirow{6}{*}{Unsupervised}} & \multicolumn{1}{c|}{PTGAN\cite{wei2018person}}                             & \multicolumn{1}{c|}{11.8}   \\ \cline{2-3} \cline{5-7} \cline{10-11} 
\multicolumn{1}{|c|}{}                               & \multicolumn{1}{c|}{UTAL\cite{Li2019Unsupervised_tpami}}                 & \multicolumn{1}{c|}{69.2}   & \multicolumn{1}{c|}{} & \multicolumn{1}{c|}{\multirow{4}{*}{Weakly}}       & \multicolumn{1}{c|}{\textbf{W-Baseline}}                    & \multicolumn{1}{c|}{61.0}   & \multicolumn{1}{c|}{} & \multicolumn{1}{c|}{}                              & \multicolumn{1}{c|}{SSG\cite{Fu2019Self_iccv}}                             & \multicolumn{1}{c|}{41.6}   \\ \cline{2-3} \cline{6-7} \cline{10-11} 
\multicolumn{1}{|c|}{}                               & \multicolumn{1}{c|}{UDA\cite{song2018unsupervised}}                & \multicolumn{1}{c|}{75.8}   & \multicolumn{1}{c|}{} & \multicolumn{1}{c|}{}                              & \multicolumn{1}{c|}{W-Local CNN}                               & \multicolumn{1}{c|}{67.6}       & \multicolumn{1}{c|}{} & \multicolumn{1}{c|}{}                              & \multicolumn{1}{c|}{TAUDL\cite{Li2018Unsupervised_eccv}}                   & \multicolumn{1}{c|}{28.4}   \\ \cline{2-3} \cline{6-7} \cline{10-11} 
\multicolumn{1}{|c|}{}                               & \multicolumn{1}{c|}{MAR\cite{yu2019unsupervised}}                  & \multicolumn{1}{c|}{67.7}   & \multicolumn{1}{c|}{} & \multicolumn{1}{c|}{}                              & \multicolumn{1}{c|}{W-MGN}                                           & \multicolumn{1}{c|}{69.8}       & \multicolumn{1}{c|}{} & \multicolumn{1}{c|}{}                              & \multicolumn{1}{c|}{UTAL\cite{Li2019Unsupervised_tpami}}                   & \multicolumn{1}{c|}{31.4}   \\ \cline{2-3} \cline{6-7} \cline{10-11} 
\multicolumn{1}{|c|}{}                               & \multicolumn{1}{c|}{DECAMEL\cite{yu2019pami}}                      & \multicolumn{1}{c|}{60.2}   & \multicolumn{1}{c|}{} & \multicolumn{1}{c|}{}                              & \multicolumn{1}{c|}{W-MGN w/o W-Tri}                          & \multicolumn{1}{c|}{67.2}       & \multicolumn{1}{c|}{} & \multicolumn{1}{c|}{}                              & \multicolumn{1}{c|}{ECN\cite{Zhong2019Invariance_cvpr}}                    & \multicolumn{1}{c|}{30.2}   \\ \cline{2-3} \cline{5-7} \cline{10-11} 
\multicolumn{1}{|c|}{}                               & \multicolumn{1}{c|}{ECN\cite{Zhong2019Invariance_cvpr}}                 & \multicolumn{1}{c|}{75.1}   &                       &                                                    &                                                                    &                             & \multicolumn{1}{c|}{} & \multicolumn{1}{c|}{}                              & \multicolumn{1}{c|}{UGA\cite{Wu2019Unsupervised_cvpr}}                     & \multicolumn{1}{c|}{49.5}   \\ \cline{2-3} \cline{9-11} 
\multicolumn{1}{|c|}{}                               & \multicolumn{1}{c|}{PAUL\cite{Yang_2019_CVPR}}                     & \multicolumn{1}{c|}{68.5}   &                       & \multicolumn{3}{c}{(c) PRID2011 test set}                                                                                                             & \multicolumn{1}{c|}{} & \multicolumn{1}{c|}{\multirow{3}{*}{Weakly}}       & \multicolumn{1}{c|}{W-MGN}                            & \multicolumn{1}{c|}{81.1}       \\ \cline{2-3} \cline{5-7} \cline{10-11} 
\multicolumn{1}{|c|}{}                               & \multicolumn{1}{c|}{HHL\cite{zhong2018generalizing}}               & \multicolumn{1}{c|}{62.2}   & \multicolumn{1}{c|}{} & \multicolumn{1}{c|}{Supervision}                   & \multicolumn{1}{c|}{Method}                                        & \multicolumn{1}{c|}{Rank-1} & \multicolumn{1}{c|}{} & \multicolumn{1}{c|}{}                              & \multicolumn{1}{c|}{W-Local CNN}                                       & \multicolumn{1}{c|}{80.6}       \\ \cline{2-3} \cline{5-7} \cline{10-11} 
\multicolumn{1}{|c|}{}                               & \multicolumn{1}{c|}{Distilled\cite{Wu_2019_CVPR}}                  & \multicolumn{1}{c|}{61.5}   & \multicolumn{1}{c|}{} & \multicolumn{1}{c|}{\multirow{9}{*}{Fully}}        & \multicolumn{1}{c|}{KISSME \cite{koestinger2012large}}             & \multicolumn{1}{c|}{18.2}   & \multicolumn{1}{c|}{} & \multicolumn{1}{c|}{}                              & \multicolumn{1}{c|}{W-MGN w/o W-Tri}                                                   & \multicolumn{1}{c|}{77.9}       \\ \cline{2-3} \cline{6-7} \cline{9-11} 
\multicolumn{1}{|c|}{}                               & \multicolumn{1}{c|}{Smooting\cite{Wang2020Smoothing_cvpr}}                 & \multicolumn{1}{c|}{83.0}   & \multicolumn{1}{c|}{} & \multicolumn{1}{c|}{}                              & \multicolumn{1}{c|}{MAHAL}                                         & \multicolumn{1}{c|}{16}     & \multicolumn{1}{c}{} &                        &                                  &      \\  \cline{1-3} \cline{6-7} 
\multicolumn{1}{|c|}{\multirow{5}{*}{Semi}}          & \multicolumn{1}{c|}{SPACO\cite{ma2017self-paced}}                  & \multicolumn{1}{c|}{68.3}   & \multicolumn{1}{c|}{} & \multicolumn{1}{c|}{}                              & \multicolumn{1}{c|}{L2}                                            & \multicolumn{1}{c|}{25}     &                       &                     \multicolumn{3}{c}{(e) SYSU-$30k$} \\ \cline{2-3} \cline{6-7}\cline{9-11}
\multicolumn{1}{|c|}{}                               & \multicolumn{1}{c|}{HHL\cite{zhong2018generalizing}}               & \multicolumn{1}{c|}{54.4}   & \multicolumn{1}{c|}{} & \multicolumn{1}{c|}{}                              & \multicolumn{1}{c|}{XQDA \cite{liao2015person}}                    & \multicolumn{1}{c|}{39}     &                       & \multicolumn{1}{|c|}{Supervision}                   & \multicolumn{1}{c|}{Method}                                                & \multicolumn{1}{c|}{Rank-1} \\ \cline{2-3} \cline{6-7} \cline{9-11} 
\multicolumn{1}{|c|}{}                               & \multicolumn{1}{c|}{Distilled\cite{Wu_2019_CVPR}}                  & \multicolumn{1}{c|}{63.9}   & \multicolumn{1}{c|}{} & \multicolumn{1}{c|}{}                              & \multicolumn{1}{c|}{P2SNet \cite{wang2017p2snet}}                  & \multicolumn{1}{c|}{60.5}   & \multicolumn{1}{c|}{} & \multicolumn{1}{c|}{\multirow{6}{*}{Fully}}                    & \multicolumn{1}{c|}{\ddag DARI \cite{wang2016dari}}                & \multicolumn{1}{c|}{11.2}   \\ \cline{2-3} \cline{6-7} \cline{10-11} 
\multicolumn{1}{|c|}{}                               & \multicolumn{1}{c|}{One Example\cite{wu2019progressive}}           & \multicolumn{1}{c|}{70.1}   & \multicolumn{1}{c|}{} & \multicolumn{1}{c|}{}                              & \multicolumn{1}{c|}{\textbf{Baseline}}                     & \multicolumn{1}{c|}{71.8}   & \multicolumn{1}{c|}{} &      & \multicolumn{1}{|c|}{\ddag DF \cite{ding2015deep}}                  & \multicolumn{1}{c|}{10.3}   \\ \cline{2-3} \cline{6-7} \cline{10-11} 
\multicolumn{1}{|c|}{}                               & \multicolumn{1}{c|}{Many Examples\cite{wu2019progressive}}         & \multicolumn{1}{c|}{82.5}   & \multicolumn{1}{c|}{} & \multicolumn{1}{c|}{}                              & \multicolumn{1}{c|}{MGN \cite{Wang2018Wang_mm} }                                           & \multicolumn{1}{c|}{74.6}       & \multicolumn{1}{c|}{} & \multicolumn{1}{c|}{}                              &  \multicolumn{1}{c|}{\ddag\textbf{Baseline}}            & \multicolumn{1}{c|}{20.1}   \\ \cline{1-3} \cline{6-7} \cline{10-11} 
\multicolumn{1}{|c|}{\multirow{6}{*}{Weakly}}        & \multicolumn{1}{c|}{\textbf{W-Baseline}}                    & \multicolumn{1}{c|}{88.6}   & \multicolumn{1}{c|}{} & \multicolumn{1}{c|}{}                              & \multicolumn{1}{c|}{MGN w/o Tri}                          & \multicolumn{1}{c|}{72.5}       & \multicolumn{1}{c|}{}                              &\multicolumn{1}{c|}{}                              & \multicolumn{1}{c|}{\ddag Local CNN \cite{Yang2018Local_mm}}                                       & \multicolumn{1}{c|}{23.0}       \\ \cline{2-3} \cline{6-7} \cline{10-11} 
\multicolumn{1}{|c|}{}                               & \multicolumn{1}{c|}{W-Local CNN}                               & \multicolumn{1}{c|}{95.7}       & \multicolumn{1}{c|}{} & \multicolumn{1}{c|}{}                              & \multicolumn{1}{c|}{Local CNN \cite{Yang2018Local_mm}}                               & \multicolumn{1}{c|}{74.2}       & \multicolumn{1}{c|}{} & \multicolumn{1}{c|}{}                              & \multicolumn{1}{c|}{\ddag MGN \cite{Wang2018Wang_mm}}                                                   & \multicolumn{1}{c|}{23.6}       \\ \cline{2-3} \cline{5-7} \cline{10-11} 
\multicolumn{1}{|c|}{}                               & \multicolumn{1}{c|}{W-MGN}                                           & \multicolumn{1}{c|}{95.5}       & \multicolumn{1}{c|}{} & \multicolumn{1}{c|}{\multirow{4}{*}{Weakly}}       & \multicolumn{1}{c|}{W-Local CNN}                               & \multicolumn{1}{c|}{71.6}       & \multicolumn{1}{c|}{} & \multicolumn{1}{c|}{}                              &\multicolumn{1}{c|}{\ddag MGN w/o Tri}                                  & \multicolumn{1}{c|}{21.5}       \\ \cline{2-3} \cline{6-7} \cline{9-11} 
\multicolumn{1}{|c|}{}                               & \multicolumn{1}{c|}{W-MGN w/o W-Tri}                          & \multicolumn{1}{c|}{92.9}       & \multicolumn{1}{c|}{} & \multicolumn{1}{c|}{}                              & \multicolumn{1}{c|}{\textbf{W-Baseline}}                    & \multicolumn{1}{c|}{68.0}     & \multicolumn{1}{c|}{} & \multicolumn{1}{c|}{}                              & \multicolumn{1}{c|}{\textbf{W-Baseline}}                            & \multicolumn{1}{c|}{26.9}   \\ \cline{2-3}\cline{6-7} \cline{10-11} 
\multicolumn{1}{|c|}{}         &                    \multicolumn{1}{c|}{W-MGN w/o graph}                                                 &       \multicolumn{1}{c|}{88.4}                      & \multicolumn{1}{c|}{} & \multicolumn{1}{c|}{}                              & \multicolumn{1}{c|}{W-MGN}                                           & \multicolumn{1}{c|}{72.7}       & \multicolumn{1}{c|}{} & \multicolumn{1}{c|}{\multirow{4}{*}{Weakly}}       & \multicolumn{1}{c|}{W-Local CNN}                                       & \multicolumn{1}{c|}{28.8}       \\\cline{2-3} \cline{6-7} \cline{10-11} 
\multicolumn{1}{|c|}{}                                     &                  \multicolumn{1}{|c|}{W-MGN w/o pair}                                                 &              \multicolumn{1}{|c|}{94.0}               & \multicolumn{1}{c|}{} & \multicolumn{1}{c|}{}                              & \multicolumn{1}{c|}{W-MGN w/o W-Tri}                          & \multicolumn{1}{c|}{70.7}       & \multicolumn{1}{c|}{} & \multicolumn{1}{c|}{}                              & \multicolumn{1}{c|}{W-MGN}                                                   & \multicolumn{1}{c|}{29.5}       \\ \cline{1-3}\cline{5-7}\cline{10-11} 
                                                     &                                                                    &                             &                       &                                                    &                                                                    &                             & \multicolumn{1}{c|}{} & \multicolumn{1}{c|}{}                              &  \multicolumn{1}{c|}{W-MGN w/o W-Tri}                                  & \multicolumn{1}{c|}{26.7}       \\ \cline{9-11} 
\end{tabular}\label{tab:cmp_sota}
\vspace{-11pt}
\end{table*}

\subsubsection{Compatibility with fully supervised learning tricks}
Intuitively, a weakly supervised Re-ID problem is likely to be upper bounded by fully supervised learning with all annotations. Next, we investigate the accuracy of our approach with respect to models with different fully-supervised learning capacities. Experiments are conducted on PRID2011 and CUHK03.

{We first evaluate two different fully supervised learning baseline models with and without re-ranking post-process. Then, we evaluate them in the weakly supervised learning scenario. The setting is similar to the aforementioned fully supervised learning, except that all of the image-level annotations are replaced with bag-level annotations in the training set. 
}

{It is observed that our rank-1 accuracy with weak annotations is close to that of using strong annotations, as shown in Table \ref{tab:ablation} (c)-(d). Moveover, weakly supervised learning with a stronger baseline (`weakly supervised \textbf{+ RK}') yields better performance. For example, in the weak annotation setting, ``weakly supervised \textbf{+ RK}'' yields 68.0\% on PRID2011, compared to 39.9\% obtained by ``weakly supervised'', a relative improvement of 70.4\%. This comparison verifies the compatibility of our method with existing frameworks; namely, existing tricks (e.g., re-ranking) for fully supervised learning could also be applied to the weakly supervised Re-ID.}

\subsection{Comparison with the State-of-the-Arts} \label{sect:sota}
In this section, we compare our weakly supervised approach with the best-performing fully-supervised / semi-supervised / unsupervised methods.

\subsubsection{Accuracy on Market-1501}\label{sect:market}
Our weakly supervised Re-ID is compared with state-of-the-art fully-supervised /unsupervised / semi-supervised methods. 

{\textbf{Fully supervised Re-ID.} We compare our method with the fully supervised Re-ID models. Fifteen representative state-of-the-art methods are used as comparison methods, including MSCAN \cite{li2017learning}, DF \cite{zhao2017deeply}, SSM \cite{bai2017scalable}, SVDNet \cite{sun2017svdnet}, GAN \cite{zheng2017unlabeled}, PDF \cite{Su2017Pose_iccv}, TriNet \cite{hermans2017defense}, TriNet + Era. + reranking \cite{zhong2017random}, PCB \cite{sun2018beyond}, VPM \cite{sun2019perceive}, JDGL \cite{zheng2019joint}, AANet \cite{Tay_2019_CVPR}, Local CNN \cite{Yang2018Local_mm}, and MGN \cite{Wang2018Wang_mm}. Comparison results are provided in Table \ref{tab:cmp_sota} (a). Our approach achieves very competitive accuracy. For example, our W-MGN and W-Local CNN achieve respectively a rank-1 accuracy of 95.5\% and 95.7\%, which surpass many of the compared fully-supervised methods. These results verify the effectiveness of our method.}

To validate the superiority of our weakly supervised Re-ID over previous annotation-saving Re-ID works, we further compare our method with state-of-the-art unsupervised and semi-supervised Re-ID methods.

\textbf{Unsupervised Re-ID.} In Table \ref{tab:cmp_sota} (a), we compare our method with 11 current best-performing models for unsupervised Re-ID, including CAMEL \cite{yu2017cross}, TAUDL\cite{li2018unsupervised}, UTAL \cite{Li2019Unsupervised_tpami}, UDA\cite{song2018unsupervised}, MAR\cite{yu2019unsupervised}, DECAMEL\cite{yu2019pami}, ECN\cite{Zhong2019Invariance_cvpr}, PAUL\cite{Yang_2019_CVPR}, HHL\cite{zhong2018generalizing}, Distilled\cite{Wu_2019_CVPR}, and Smooting \cite{Wang2020Smoothing_cvpr}. The results in Table \ref{tab:cmp_sota} (a) show that our weakly supervised Re-ID has obtained significant gain over unsupervised Re-ID methods. For instance, our W-MGN outperforms the best-performing model UDA\cite{song2018unsupervised} by a large margin (i.e., 19.7\%). Note that compared to unsupervised Re-ID, the annotation effort of our weakly supervised Re-ID is also very inexpensive. These results verify the effectiveness of our method again.

\textbf{Semi-supervised Re-ID.}
In Table \ref{tab:cmp_sota} (a), we compare our method with the semi-supervised Re-ID models. Five representative state-of-the-art methods are used as competing methods, including SPACO \cite{ma2017self-paced}, HHL \cite{zhong2018generalizing}, Distilled\cite{Wu_2019_CVPR}, One Example\cite{wu2019progressive}, and Many Examples\cite{wu2019progressive}. The results show that our weakly supervised Re-ID problem has obtained significant gain over semi-supervised Re-ID methods. For instance, our method outperforms the best-performing model ``ManyExamples'' \cite{wu2019progressive} by a large margin (i.e., 13\%), indicating that as an annotation-saving method, our weakly Re-ID obtains higher accuracy than semi-supervised Re-ID.

\subsubsection{Accuracy on CUHK03} \label{sect:cuhk03}
Our weakly supervised Re-ID is compared with state-of-the-art methods in two groups on CUHK03, including the traditional fully-supervised Re-ID and the unsupervised Re-ID. The semi-supervised Re-ID is not compared here because existing works on semi-supervised Re-ID do not provide results on this dataset.

\textbf{Fully supervised Re-ID.} In Table \ref{tab:cmp_sota} (b), we compare our method with the eleven current best models, including BOW+XQDA \cite{zheng2015scalable}, PUL \cite{fan2018unsupervised}, LOMO+XQDA \cite{liao2015person}, IDE(R) \cite{zheng2016person}, IDE+DaF \cite{yu2017divide}, IDE+XQ+reranking \cite{zhong2017re}, PAN, DPFL \cite{chen2017person}, and newly proposed methods such as SVDNet \cite{sun2017svdnet}, TriNets \cite{zhong2017random}, ACNet\cite{wang2019adaptively}, Local CNN \cite{Yang2018Local_mm}, MGN \cite{Wang2018Wang_mm}, and TreeConv \cite{Wang2020Grammatically_kdd}. Our W-MGN and W-Local CNN achieve a rank-1 accuracy of 69.8\% and 67.6\%, which surpass many of the compared fully-supervised methods. These results verify the effectiveness of our method.

\textbf{Unsupervised Re-ID.} In Table \ref{tab:cmp_sota} (b), we compare our method with the unsupervised Re-ID models. Three representative state-of-the-art methods are used as competing methods, including CAMEL \cite{yu2017cross}, PatchNet \cite{Yang_2019_CVPR}, and PAUL \cite{Yang_2019_CVPR}. The results in Table \ref{tab:cmp_sota} (b) show that our weakly supervised Re-ID problem has obtained significant gain over unsupervised Re-ID methods. For instance, our W-MGN outperforms the best-performing model PAUL\cite{Yang_2019_CVPR} by a large margin (i.e., 17.5\%). Given that our method also saves the annotation effort, we believe our weakly-supervised Re-ID balances well between annotation and accuracy.

\subsubsection{Accuracy on MSMT17} \label{sect:cuhk03}
{Our weakly supervised Re-ID is compared with state-of-the-art methods fully-supervised / unsupervised Re-ID on MSMT17. Semi-supervised methods are not compared here because previous works on semi-supervised Re-ID do not provide results on this dataset.}

{\textbf{Fully supervised Re-ID.} We compare our method with the eleven current best models, including PDC\cite{Su2017Pose_iccv}, GLAD\cite{Wei2017GLAD_acmmm}, ABD-Net\cite{Chen2019ABD_iccv}, GoogleNet\cite{Szegedy2015Going_cvpr}, IANet\cite{Hou2019Interaction_cvpr}, SFT\cite{Luo2019Spectral_iccv}, Auto-ReID\cite{Quan2019Auto_iccv}, MVP\cite{Chen2019MVP_iccv}, Verif-Identif\cite{Zheng2018Discriminatively_tomccap}, PCB\cite{sun2018beyond}, JDGL\cite{zheng2019joint}, ShuffleNet\cite{Zhang2018ShuffleNet_cvpr}, MobileNetV2\cite{Sandler2018MobileNetV2_cvpr}, OSNet\cite{Zhou2019Omni_iccv}, Local CNN \cite{Yang2018Local_mm}, and MGN \cite{Wang2018Wang_mm}. Comparison results are provided in Table \ref{tab:cmp_sota} (d). Our W-MGN and W-Local CNN achieve a rank-1 accuracy of 81.1\% and 80.6\%, which surpass many of the compared fully-supervised methods. These results verify the effectiveness of our method.}

{\textbf{Unsupervised Re-ID.} We compare our method with the unsupervised Re-ID models, as shown in Table \ref{tab:cmp_sota} (b). Five representative state-of-the-art methods are used as competing methods, including PTGAN\cite{wei2018person}, SSG\cite{Fu2019Self_iccv}, TAUDL\cite{Li2018Unsupervised_eccv}, UTAL\cite{Li2019Unsupervised_tpami}, ECN\cite{Zhong2019Invariance_cvpr}, and UGA\cite{Wu2019Unsupervised_cvpr}. The results in Table \ref{tab:cmp_sota} (d) show that our weakly supervised Re-ID problem has obtained significant gain over unsupervised Re-ID methods. For instance, our W-MGN outperforms the best-performing model UGA\cite{Wu2019Unsupervised_cvpr} by a large margin (i.e., 81.1\% \emph{vs.} 49.5\%). Given that our method also saves the annotation effort, we believe our weakly supervised Re-ID balances well between annotation and accuracy.}

\subsubsection{Accuracy on PRID2011}
In Table \ref{tab:cmp_sota} (c) and Fig. \ref{fig:cmp_sota} (a), we compare the results of our model with five current best models: the KISSME distance learning method \cite{koestinger2012large}, MAHAL, L2, and XQDA \cite{liao2015person}, P2SNet \cite{wang2017p2snet}, Local CNN \cite{Yang2018Local_mm}, and MGN \cite{Wang2018Wang_mm}. For KISSME, MAHAL, L2, and XQDA, deep features \cite{zheng2016mars} are utilized to represent an image of a person. For P2SNet, we train the model based on the image-to-video setting but sample one frame from each video to formulate the image-to-image setting. Our W-MGN and W-Local CNN achieve a rank-1 accuracy of 72.7\% and 71.6\%, which surpass many of the compared fully-supervised methods. These results verify the effectiveness of our method.

\begin{figure}[t]
 \begin{center}
\includegraphics[width=1.0\linewidth]{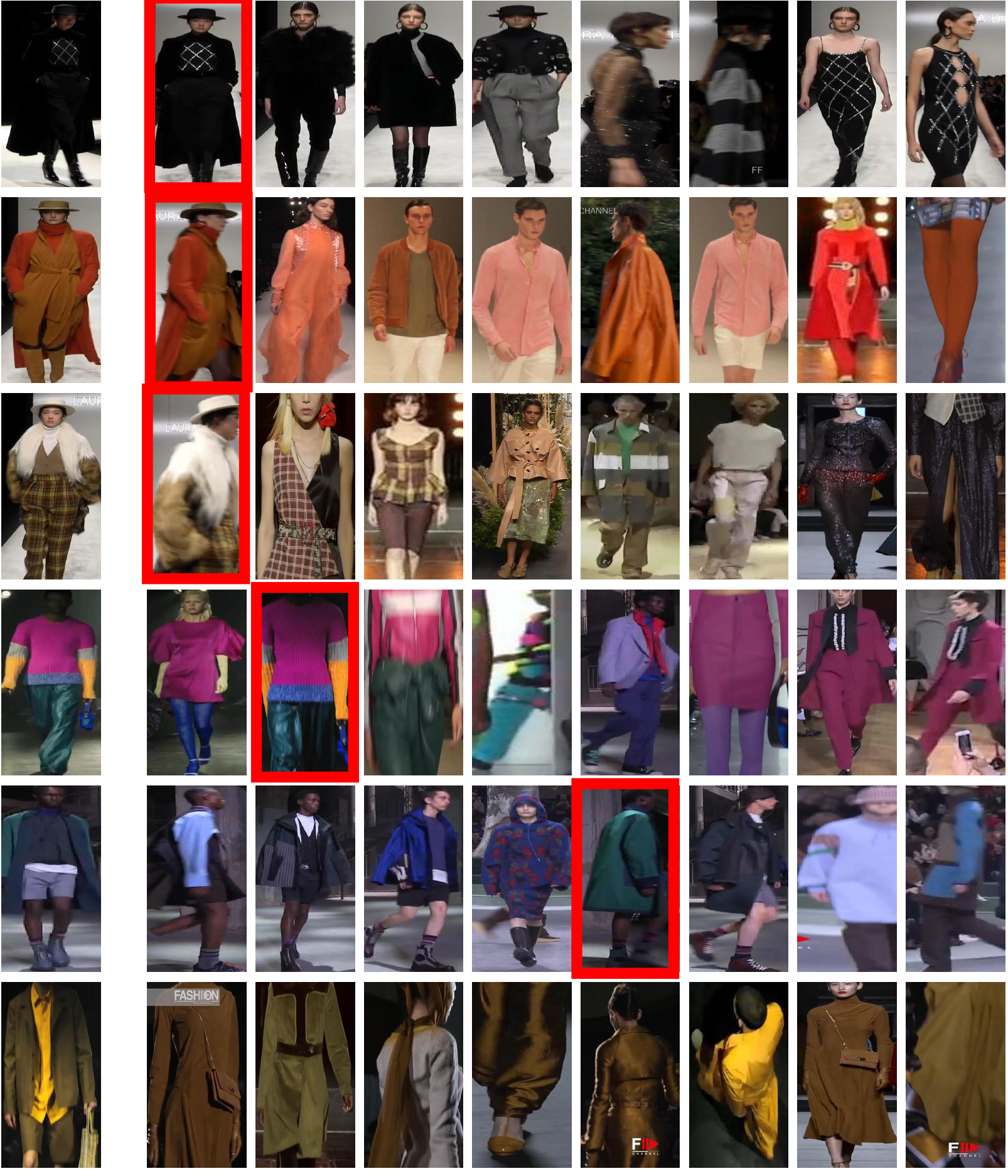}
\end{center}
\vspace{-11pt}
\caption{\small{{Search examples of W-MGN on SYSU-30$k$ dataset. Each row represents a ranking result with the first image being the query and the rest images being the returned list. The image with the red bounding box is the matched one.}}}\label{fig:search}
\vspace{-11pt}
\end{figure}

\subsubsection{Accuracy on SYSU-{{{$30k$}}}}
As SYSU-{$30k$} is the only weakly supervised Re-ID dataset and our method is the only weakly supervised Re-ID method, we propose to compare the traditional fully supervised Re-ID models with our weakly supervised method by using transfer learning. Specifically, six representative fully supervised Re-ID models including DARI \cite{wang2016dari}, DF \cite{ding2015deep}, TriNet\cite{zhong2017random}, Local CNN \cite{Yang2018Local_mm}, MGN \cite{Wang2018Wang_mm}, and MGN without triplet loss are first trained on CUHK03. Then, they are used to performed cross-dataset evaluation on the test set of SYSU-{$30k$}. In contrast, our weakly-supervised Re-ID is trained on the training set of the SYSU-30$k$ with weak annotations and then is tested on the test set of SYSU-30$k$.

{Table \ref{tab:cmp_sota} (e) and Fig. \ref{fig:cmp_sota} (b) are the results of the comparisons. It is observed that our W-MGN achieves state-of-the-art performance (29.5\%), even though it is trained in a weakly supervised manner while the comparison methods are trained with full supervision. The success may be attributed to two reasons. First, our model is quite effective due to the graphical modeling that generates reliable pseudo labels as compensation for the absence of strong labels. Second, the large-scale SYSU-30$k$ dataset provides rich knowledge that improves the capacity of our model, even though SYSU-30$k$ is annotated weakly.} 

{We also qualitatively present some query examples of W-MGN for the SYSU-30$k$ dataset in Fig. \ref{fig:search}. 
Each row represents a ranking result with the first image being the query image and the rest being the result list. The matched one in the returned list is highlighted by a red bounding box. This figure exhibits the difficulty of this dataset. Actually, in the failed examples, the images ranked higher than the matched one often look more closer to the query image as in Row 4–6.}

\begin{table}
\caption{\small{Computational complexity of weakly and fully supervised Re-ID. \textbf{secs / 100 images:} the time of forward-passing 100 images in the testing stage or the cycle of a forward-backward passing in the training stage when the batch size is 100.}}
\label{tab:cmp_cost}
\centering
\begin{tabular}{|c|c|c|cc}
\hline
& weakly (secs / 100 images) & fully (secs / 100 image)\\
\hline
 Testing   & 0.0559 & 0.0559 \\
 Training  & 0.2453 & 0.2448 \\
\hline
\end{tabular}
\vspace{-11pt}
\end{table}

\subsubsection{Computational Complexity}
Table \ref{tab:cmp_cost} compares the computational time of Re-ID in the context of weak supervision to that in the context of full supervision in terms of time cost per 100 images. For a fair comparison, both methods are individually trained on the same desktop with 1 Titan-x GPU. As shown in the table, the weakly and fully supervised Re-ID methods have similar computational costs. Specifically, in the testing phase, both methods share the same computational costs. Even in the training phase, our method only performs $0.002\times$ slower than the fully supervised Re-ID (0.2453 vs. 0.2448 seconds per 100 images using TITAN X.).

\section{Conclusion}\label{sect:conclusion}

We have considered a more realistic Re-ID problem challenge: the weakly supervised Re-ID problem. To address this new problem, we proposed a graphical model to capture the dependencies among images in each weakly annotated bag. We further propose a weakly annotated Re-ID dataset (i.e., SYSU-$30k$) to facilitate future research, which is currently the largest Re-ID benchmark. Extensive experiments have conducted on our SYSU-$30k$ dataset and other four public Re-ID datasets and a superior performance is achieved with our proposed model, providing a promising and appealing conclusion that learning a Re-ID model with less annotation efforts is possible and feasible.  
Future work will include building automated models \cite{Li2020Block_cvpr} for the weakly supervised Re-ID .

\section*{Acknowledgement}
This work was supported in part by the State Key Development Program under Grant 2018YFC0830103, in part by the National Natural Science Foundation of China under Grant 61876224, U1811463, 61622214, and 61836012, and in part by Guangdong Natural Science Foundation under Grant 2017A030312006.

%------------------------------------------------------------------------
\bibliographystyle{IEEEtran}
\bibliography{wsreid}

\begin{thebibliography}{10}
\providecommand{\url}[1]{#1}
\csname url@samestyle\endcsname
\providecommand{\newblock}{\relax}
\providecommand{\bibinfo}[2]{#2}
\providecommand{\BIBentrySTDinterwordspacing}{\spaceskip=0pt\relax}
\providecommand{\BIBentryALTinterwordstretchfactor}{4}
\providecommand{\BIBentryALTinterwordspacing}{\spaceskip=\fontdimen2\font plus
\BIBentryALTinterwordstretchfactor\fontdimen3\font minus
  \fontdimen4\font\relax}
\providecommand{\BIBforeignlanguage}[2]{{%
\expandafter\ifx\csname l@#1\endcsname\relax
\typeout{** WARNING: IEEEtran.bst: No hyphenation pattern has been}%
\typeout{** loaded for the language `#1'. Using the pattern for}%
\typeout{** the default language instead.}%
\else
\language=\csname l@#1\endcsname
\fi
#2}}
\providecommand{\BIBdecl}{\relax}
\BIBdecl

\bibitem{li2014deepreid}
W.~Li, R.~Zhao, T.~Xiao, and X.~Wang, ``Deepreid: Deep filter pairing neural
  network for person re-identification,'' in \emph{{IEEE} Conference on
  Computer Vision and Pattern Recognition}, 2014, pp. 152--159.

\bibitem{zheng2015scalable}
L.~Zheng, L.~Shen, L.~Tian, S.~Wang, J.~Wang, and Q.~Tian, ``Scalable person
  re-identification: A benchmark,'' in \emph{{IEEE} International Conference on
  Computer Vision}.\hskip 1em plus 0.5em minus 0.4em\relax IEEE, 2015, pp.
  1116--1124.

\bibitem{gray2008viewpoint}
D.~Gray and H.~Tao, ``Viewpoint invariant pedestrian recognition with an
  ensemble of localized features,'' in \emph{Proceedings of the European
  Conference on Computer Vision}, 2008, pp. 262--275.

\bibitem{Lafferty2001Conditional_icml}
J.~D. Lafferty, A.~McCallum, and F.~C.~N. Pereira, ``Conditional random fields:
  Probabilistic models for segmenting and labeling sequence data,'' in
  \emph{Proceedings of the International Conference on Machine Learning}, 2001,
  pp. 282--289.

\bibitem{koestinger2012large}
M.~Koestinger, M.~Hirzer, P.~Wohlhart, P.~M. Roth, and H.~Bischof, ``Large
  scale metric learning from equivalence constraints,'' in \emph{{IEEE}
  Conference on Computer Vision and Pattern Recognition}, 2012, pp. 2288--2295.

\bibitem{ding2015deep}
S.~Ding, L.~Lin, G.~Wang, and H.~Chao, ``Deep feature learning with relative
  distance comparison for person re-identification,'' \emph{Pattern
  Recognition}, vol.~48, no.~10, pp. 2993--3003, 2015.

\bibitem{wang2016dari}
G.~Wang, L.~Lin, S.~Ding, Y.~Li, and Q.~Wang, ``Dari: Distance metric and
  representation integration for person verification.'' in \emph{Proceedings of
  The AAAI Conference on Artificial Intelligence}, 2016, pp. 3611--3617.

\bibitem{lin2017cross}
L.~Lin, G.~Wang, W.~Zuo, X.~Feng, and L.~Zhang, ``Cross-domain visual matching
  via generalized similarity measure and feature learning,'' \emph{{IEEE}
  Transactions on Pattern Analysis and Machine Intelligence}, vol.~39, no.~6,
  pp. 1089--1102, 2017.

\bibitem{wang2017p2snet}
G.~Wang, J.~Lai, and X.~Xie, ``P2snet: Can an image match a video for person
  re-identification in an end-to-end way?'' \emph{IEEE Transactions on Circuits
  and Systems for Video Technology}, pp. 2777--2787, 2017.

\bibitem{guangcong2019aaai}
G.~Wang, J.~Lai, P.~Huang, and X.~Xie, ``Spatial-temporal person
  re-identification,'' in \emph{Proceedings of The AAAI Conference on
  Artificial Intelligence}, 2019.

\bibitem{zhuo2018occluded}
J.~Zhuo, Z.~Chen, J.~Lai, and G.~Wang, ``Occluded person re-identification,''
  in \emph{Proceedings of the IEEE International Conference on Multimedia and
  Expo}, 2018, pp. 1--6.

\bibitem{li2017person}
S.~Li, T.~Xiao, H.~Li, B.~Zhou, D.~Yue, and X.~Wang, ``Person search with
  natural language description,'' in \emph{{IEEE} Conference on Computer Vision
  and Pattern Recognition}, 2017, pp. 1970--1979.

\bibitem{Wang2020Transferable_cvpr}
H.~Wang, G.~Wang, Y.~Li, D.~Zhang, and L.~Lin, ``Transferable, controllable,
  and inconspicuous adversarial attacks on person re-identification with deep
  mis-ranking,'' in \emph{{IEEE} Conference on Computer Vision and Pattern
  Recognition}, 2020.

\bibitem{yu2017cross}
H.-X. Yu, A.~Wu, and W.-S. Zheng, ``Cross-view asymmetric metric learning for
  unsupervised person re-identification,'' in \emph{{IEEE} International
  Conference on Computer Vision}, 2017, pp. 994--1002.

\bibitem{liang2018m2m}
W.~Liang, G.~Wang, J.~Lai, and J.~Zhu, ``{M2M-GAN:} many-to-many generative
  adversarial transfer learning for person re-identification,'' \emph{CoRR},
  vol. abs/1811.03768, 2018.

\bibitem{song2018unsupervised}
L.~Song, C.~Wang, L.~Zhang, B.~Du, Q.~Zhang, C.~Huang, and X.~Wang,
  ``Unsupervised domain adaptive re-identification: Theory and practice,''
  \emph{CoRR}, vol. abs/1807.11334, 2018.

\bibitem{yu2019unsupervised}
H.~Yu, W.~Zheng, A.~Wu, X.~Guo, S.~Gong, and J.~Lai, ``Unsupervised person
  re-identification by soft multilabel learning,'' in \emph{{IEEE} Conference
  on Computer Vision and Pattern Recognition}, 2019, pp. 2148--2157.

\bibitem{fan2018unsupervised}
H.~Fan, L.~Zheng, C.~Yan, and Y.~Yang, ``Unsupervised person re-identification:
  Clustering and fine-tuning,'' \emph{{ACM} Transactions on Multimedia
  Computing, Communications, and Applications}, vol.~14, no.~4, p.~83, 2018.

\bibitem{Li2019Unsupervised_tpami}
M.~Li, X.~Zhu, and S.~Gong, ``Unsupervised tracklet person re-identification,''
  \emph{{IEEE} Transactions on Pattern Analysis and Machine Intelligence}, vol.
  pages 1–1, 2019.

\bibitem{wu2019progressive}
Y.~Wu, Y.~Lin, X.~Dong, Y.~Yan, W.~Bian, and Y.~Yang, ``Progressive learning
  for person re-identification with one example,'' \emph{{IEEE} Transactions on
  Image Processing}, vol.~28, no.~6, pp. 2872--2881, 2019.

\bibitem{wu2018exploit}
Y.~Wu, Y.~Lin, X.~Dong, Y.~Yan, W.~Ouyang, and Y.~Yang, ``Exploit the unknown
  gradually: One-shot video-based person re-identification by stepwise
  learning,'' in \emph{{IEEE} Conference on Computer Vision and Pattern
  Recognition}, 2018, pp. 5177--5186.

\bibitem{meng2019weakly}
J.~Meng, S.~Wu, and W.~Zheng, ``Weakly supervised person re-identification,''
  in \emph{{IEEE} Conference on Computer Vision and Pattern Recognition}, 2019,
  pp. 760--769.

\bibitem{mahajan2018exploring}
D.~Mahajan, R.~B. Girshick, V.~Ramanathan, K.~He, M.~Paluri, Y.~Li,
  A.~Bharambe, and L.~van~der Maaten, ``Exploring the limits of weakly
  supervised pretraining,'' in \emph{Proceedings of the European Conference on
  Computer Vision}, 2018, pp. 185--201.

\bibitem{wang2017deep}
G.~Wang, X.~Xie, J.~Lai, and J.~Zhuo, ``Deep growing learning,'' in
  \emph{{IEEE} International Conference on Computer Vision}, 2017, pp.
  2812--2820.

\bibitem{wang2017learning}
G.~Wang, P.~Luo, L.~Lin, and X.~Wang, ``Learning object interactions and
  descriptions for semantic image segmentation,'' in \emph{{IEEE} Conference on
  Computer Vision and Pattern Recognition}, 2017, pp. 5859--5867.

\bibitem{luo2017deep}
P.~Luo, G.~Wang, L.~Lin, and X.~Wang, ``Deep dual learning for semantic image
  segmentation,'' in \emph{{IEEE} Conference on Computer Vision and Pattern
  Recognition}, 2017, pp. 2718--2726.

\bibitem{lin2016deep}
L.~Lin, G.~Wang, R.~Zhang, R.~Zhang, X.~Liang, and W.~Zuo, ``Deep structured
  scene parsing by learning with image descriptions,'' in \emph{{IEEE}
  Conference on Computer Vision and Pattern Recognition}, 2016, pp. 2276--2284.

\bibitem{Zhang2019Hierarchical_pami}
R.~Zhang, L.~Lin, G.~Wang, M.~Wang, and W.~Zuo, ``Hierarchical scene parsing by
  weakly supervised learning with image descriptions,'' \emph{{IEEE} Trans.
  Pattern Anal. Mach. Intell.}, vol.~41, no.~3, pp. 596--610, 2019.

\bibitem{cinbis2017weakly}
R.~G. Cinbis, J.~Verbeek, and C.~Schmid, ``Weakly supervised object
  localization with multi-fold multiple instance learning,'' \emph{{IEEE}
  Transactions on Pattern Analysis and Machine Intelligence}, vol.~39, no.~1,
  pp. 189--203, 2017.

\bibitem{chen2018deeplab}
L.-C. Chen, G.~Papandreou, I.~Kokkinos, K.~Murphy, and A.~L. Yuille, ``Deeplab:
  Semantic image segmentation with deep convolutional nets, atrous convolution,
  and fully connected crfs,'' \emph{{IEEE} Transactions on Pattern Analysis and
  Machine Intelligence}, vol.~40, no.~4, pp. 834--848, 2018.

\bibitem{wu2013constrained}
B.~Wu, Y.~Zhang, B.~Hu, and Q.~Ji, ``Constrained clustering and its application
  to face clustering in videos,'' in \emph{{IEEE} Conference on Computer Vision
  and Pattern Recognition}, 2013, pp. 3507--3514.

\bibitem{shen2018person}
Y.~Shen, H.~Li, S.~Yi, D.~Chen, and X.~Wang, ``Person re-identification with
  deep similarity-guided graph neural network,'' in \emph{Proceedings of the
  European Conference on Computer Vision}.\hskip 1em plus 0.5em minus
  0.4em\relax Springer, 2018, pp. 508--526.

\bibitem{zheng2017unlabeled}
Z.~Zheng, L.~Zheng, and Y.~Yang, ``Unlabeled samples generated by {GAN} improve
  the person re-identification baseline in vitro,'' in \emph{{IEEE}
  International Conference on Computer Vision}, 2017, pp. 3774--3782.

\bibitem{wei2018person}
L.~Wei, S.~Zhang, W.~Gao, and Q.~Tian, ``Person transfer gan to bridge domain
  gap for person re-identification,'' in \emph{{IEEE} Conference on Computer
  Vision and Pattern Recognition}.\hskip 1em plus 0.5em minus 0.4em\relax IEEE,
  2018, pp. 79--88.

\bibitem{li2012human}
W.~Li, R.~Zhao, and X.~Wang, ``Human reidentification with transferred metric
  learning,'' in \emph{Proceedings of the Asian Conference on Computer Vision},
  2012, pp. 31--44.

\bibitem{hirzer2011person}
M.~Hirzer, C.~Beleznai, P.~M. Roth, and H.~Bischof, ``Person re-identification
  by descriptive and discriminative classification,'' in \emph{Scandinavian
  conference on Image analysis}, 2011, pp. 91--102.

\bibitem{cheng2011custom}
D.~S. Cheng, M.~Cristani, M.~Stoppa, L.~Bazzani, and V.~Murino, ``Custom
  pictorial structures for re-identification.'' in \emph{Proceedings of The
  British Machine Vision Conference}, vol.~1, no.~2, 2011, p.~6.

\bibitem{redmon2017yolo9000}
J.~Redmon and A.~Farhadi, ``{YOLO9000:} better, faster, stronger,'' in
  \emph{{IEEE} Conference on Computer Vision and Pattern Recognition}, 2017,
  pp. 6517--6525.

\bibitem{sun2018beyond}
Y.~Sun, L.~Zheng, Y.~Yang, Q.~Tian, and S.~Wang, ``Beyond part models: Person
  retrieval with refined part pooling (and a strong convolutional baseline),''
  in \emph{Proceedings of the European Conference on Computer Vision}, 2018,
  pp. 480--496.

\bibitem{he2016deep}
K.~He, X.~Zhang, S.~Ren, and J.~Sun, ``Deep residual learning for image
  recognition,'' in \emph{{IEEE} Conference on Computer Vision and Pattern
  Recognition}, 2016, pp. 770--778.

\bibitem{hermans2017defense}
A.~Hermans, L.~Beyer, and B.~Leibe, ``In defense of the triplet loss for person
  re-identification,'' \emph{CoRR}, vol. abs/1703.07737, 2017.

\bibitem{ioffe2015batch}
S.~Ioffe and C.~Szegedy, ``Batch normalization: Accelerating deep network
  training by reducing internal covariate shift,'' in \emph{Proceedings of the
  International Conference on Machine Learning}, 2015, pp. 448--456.

\bibitem{nair2010rectified}
V.~Nair and G.~E. Hinton, ``Rectified linear units improve restricted boltzmann
  machines,'' in \emph{Proceedings of the International Conference on Machine
  Learning}, 2010, pp. 807--814.

\bibitem{srivastava2014dropout}
N.~Srivastava, G.~Hinton, A.~Krizhevsky, I.~Sutskever, and R.~Salakhutdinov,
  ``Dropout: A simple way to prevent neural networks from overfitting,''
  \emph{The Journal of Machine Learning Research}, vol.~15, no.~1, pp.
  1929--1958, 2014.

\bibitem{Wang2018Wang_mm}
G.~Wang, Y.~Yuan, X.~Chen, J.~Li, and X.~Zhou, ``Learning discriminative
  features with multiple granularities for person re-identification,'' in
  \emph{{ACM} Multimedia Conference on Multimedia Conference}, 2018, pp.
  274--282.

\bibitem{zhong2017re}
Z.~Zhong, L.~Zheng, D.~Cao, and S.~Li, ``Re-ranking person re-identification
  with k-reciprocal encoding,'' in \emph{{IEEE} Conference on Computer Vision
  and Pattern Recognition}, 2017, pp. 3652--3661.

\bibitem{Yang2018Local_mm}
J.~Yang, X.~Shen, X.~Tian, H.~Li, J.~Huang, and X.~Hua, ``Local convolutional
  neural networks for person re-identification,'' in \emph{{ACM} Multimedia
  Conference on Multimedia Conference}, 2018, pp. 1074--1082.

\bibitem{li2017learning}
D.~Li, X.~Chen, Z.~Zhang, and K.~Huang, ``Learning deep context-aware features
  over body and latent parts for person re-identification,'' in \emph{{IEEE}
  Conference on Computer Vision and Pattern Recognition}, 2017, pp. 384--393.

\bibitem{Su2017Pose_iccv}
C.~Su, J.~Li, S.~Zhang, J.~Xing, W.~Gao, and Q.~Tian, ``Pose-driven deep
  convolutional model for person re-identification,'' in \emph{{IEEE}
  International Conference on Computer Vision}, 2017, pp. 3980--3989.

\bibitem{zhao2017deeply}
L.~Zhao, X.~Li, Y.~Zhuang, and J.~Wang, ``Deeply-learned part-aligned
  representations for person re-identification.'' in \emph{{IEEE} International
  Conference on Computer Vision}, 2017, pp. 3239--3248.

\bibitem{Wei2017GLAD_acmmm}
L.~Wei, S.~Zhang, H.~Yao, W.~Gao, and Q.~Tian, ``{GLAD:} global-local-alignment
  descriptor for pedestrian retrieval,'' in \emph{{ACM} Multimedia Conference
  on Multimedia Conference}, 2017, pp. 420--428.

\bibitem{bai2017scalable}
S.~Bai, X.~Bai, and Q.~Tian, ``Scalable person re-identification on supervised
  smoothed manifold,'' in \emph{{IEEE} Conference on Computer Vision and
  Pattern Recognition}, 2017, pp. 2530--2539.

\bibitem{liao2015person}
S.~Liao, Y.~Hu, X.~Zhu, and S.~Z. Li, ``Person re-identification by local
  maximal occurrence representation and metric learning,'' in \emph{{IEEE}
  Conference on Computer Vision and Pattern Recognition}, 2015, pp. 2197--2206.

\bibitem{Chen2019ABD_iccv}
T.~Chen, S.~Ding, J.~Xie, Y.~Yuan, W.~Chen, Y.~Yang, Z.~Ren, and Z.~Wang,
  ``Abd-net: Attentive but diverse person re-identification,'' in \emph{{IEEE}
  International Conference on Computer Vision}, 2019.

\bibitem{sun2017svdnet}
Y.~Sun, L.~Zheng, W.~Deng, and S.~Wang, ``Svdnet for pedestrian retrieval,'' in
  \emph{{IEEE} International Conference on Computer Vision}, 2017, pp.
  3820--3828.

\bibitem{zheng2016person}
L.~Zheng, Y.~Yang, and A.~G. Hauptmann, ``Person re-identification: Past,
  present and future,'' \emph{CoRR}, vol. abs/1610.02984, 2016.

\bibitem{Szegedy2015Going_cvpr}
C.~Szegedy, W.~Liu, Y.~Jia, P.~Sermanet, S.~E. Reed, D.~Anguelov, D.~Erhan,
  V.~Vanhoucke, and A.~Rabinovich, ``Going deeper with convolutions,'' in
  \emph{{IEEE} Conference on Computer Vision and Pattern Recognition}, 2015,
  pp. 1--9.

\bibitem{yu2017divide}
R.~Yu, Z.~Zhou, S.~Bai, and X.~Bai, ``Divide and fuse: {A} re-ranking approach
  for person re-identification,'' in \emph{Proceedings of The British Machine
  Vision Conference}, 2017.

\bibitem{Hou2019Interaction_cvpr}
R.~Hou, B.~Ma, H.~Chang, X.~Gu, S.~Shan, and X.~Chen,
  ``Interaction-and-aggregation network for person re-identification,'' in
  \emph{{IEEE} Conference on Computer Vision and Pattern Recognition}, 2019,
  pp. 9317--9326.

\bibitem{Luo2019Spectral_iccv}
C.~Luo, Y.~Chen, N.~Wang, and Z.~Zhang, ``Spectral feature transformation for
  person re-identification,'' in \emph{{IEEE} International Conference on
  Computer Vision}, 2019.

\bibitem{Quan2019Auto_iccv}
R.~Quan, X.~Dong, Y.~Wu, L.~Zhu, and Y.~Yang, ``Auto-reid: Searching for a
  part-aware convnet for person re-identification,'' in \emph{{IEEE}
  International Conference on Computer Vision}, 2019.

\bibitem{zhong2017random}
Z.~Zhong, L.~Zheng, G.~Kang, S.~Li, and Y.~Yang, ``Random erasing data
  augmentation,'' \emph{CoRR}, vol. abs/1708.04896, 2017.

\bibitem{chen2017person}
Y.~Chen, X.~Zhu, and S.~Gong, ``Person re-identification by deep learning
  multi-scale representations,'' in \emph{{IEEE} Conference on Computer Vision
  and Pattern Recognition}, 2017, pp. 2590--2600.

\bibitem{Chen2019MVP_iccv}
H.~Sun, Z.~Chen, S.~Yan, and L.~Xu, ``Mvp matching: A maximum-value perfect
  matching for mining hard samples, with application to person
  re-identification,'' in \emph{{IEEE} International Conference on Computer
  Vision}, 2019.

\bibitem{Wang2020Grammatically_kdd}
G.~Wang, G.~Wang, K.~Wang, X.~Liang, and L.~Lin, ``Grammatically recognizing
  images with tree convolution,'' in \emph{Proceedings of the 26th ACM SIGKDD
  International Conference on Knowledge Discovery and Data Mining (KDD), San
  Diego, CA, USA, August 4-8, 2020. ACM}, 2020.

\bibitem{Zheng2018Discriminatively_tomccap}
Z.~Zheng, L.~Zheng, and Y.~Yang, ``A discriminatively learned cnn embedding for
  person reidentification,'' \emph{{ACM} Transactions on Multimedia Computing,
  Communications, and Applications}, vol.~14, no.~1, pp. 13:1--13:20, 2018.

\bibitem{sun2019perceive}
Y.~Sun, Q.~Xu, Y.~Li, C.~Zhang, Y.~Li, S.~Wang, and J.~Sun, ``Perceive where to
  focus: Learning visibility-aware part-level features for partial person
  re-identification,'' in \emph{{IEEE} Conference on Computer Vision and
  Pattern Recognition}, 2019, pp. 393--402.

\bibitem{zheng2019joint}
Z.~Zheng, X.~Yang, Z.~Yu, L.~Zheng, Y.~Yang, and J.~Kautz, ``Joint
  discriminative and generative learning for person re-identification,'' in
  \emph{{IEEE} Conference on Computer Vision and Pattern Recognition}, 2019,
  pp. 2138--2147.

\bibitem{wang2019adaptively}
G.~Wang, K.~Wang, and L.~Lin, ``Adaptively connected neural networks,'' in
  \emph{{IEEE} Conference on Computer Vision and Pattern Recognition}, 2019,
  pp. 1781--1790.

\bibitem{Tay_2019_CVPR}
C.~Tay, S.~Roy, and K.~Yap, ``Aanet: Attribute attention network for person
  re-identifications,'' in \emph{{IEEE} Conference on Computer Vision and
  Pattern Recognition}, 2019, pp. 7134--7143.

\bibitem{Zhang2018ShuffleNet_cvpr}
X.~Zhang, X.~Zhou, M.~Lin, and J.~Sun, ``Shufflenet: An extremely efficient
  convolutional neural network for mobile devices,'' in \emph{{IEEE} Conference
  on Computer Vision and Pattern Recognition}, 2018, pp. 6848--6856.

\bibitem{Sandler2018MobileNetV2_cvpr}
M.~Sandler, A.~G. Howard, M.~Zhu, A.~Zhmoginov, and L.~Chen, ``Mobilenetv2:
  Inverted residuals and linear bottlenecks,'' in \emph{{IEEE} Conference on
  Computer Vision and Pattern Recognition}, 2018, pp. 4510--4520.

\bibitem{Yang_2019_CVPR}
Q.~Yang, H.~Yu, A.~Wu, and W.~Zheng, ``Patch-based discriminative feature
  learning for unsupervised person re-identification,'' in \emph{{IEEE}
  Conference on Computer Vision and Pattern Recognition}, 2019, pp. 3633--3642.

\bibitem{Zhou2019Omni_iccv}
K.~Zhou, Y.~Yang, A.~Cavallaro, and T.~Xiang, ``Omni-scale feature learning for
  person re-identification,'' in \emph{{IEEE} International Conference on
  Computer Vision}, 2019.

\bibitem{li2018unsupervised}
M.~Li, X.~Zhu, and S.~Gong, ``Unsupervised person re-identification by deep
  learning tracklet association,'' in \emph{Proceedings of the European
  Conference on Computer Vision}, 2018, pp. 772--788.

\bibitem{Fu2019Self_iccv}
Y.~Fu, Y.~Wei, G.~Wang, X.~Zhou, H.~Shi, and T.~S. Huang, ``Self-similarity
  grouping: {A} simple unsupervised cross domain adaptation approach for person
  re-identification,'' in \emph{{IEEE} International Conference on Computer
  Vision}, 2019.

\bibitem{Li2018Unsupervised_eccv}
M.~Li, X.~Zhu, and S.~Gong, ``Unsupervised person re-identification by deep
  learning tracklet association,'' in \emph{Proceedings of the European
  Conference on Computer Vision}, 2018, pp. 772--788.

\bibitem{yu2019pami}
H.~Yu, A.~Wu, and W.~Zheng, ``Unsupervised person re-identification by deep
  asymmetric metric embedding,'' \emph{{IEEE} Transactions on Pattern Analysis
  and Machine Intelligence}, pp. 1--1, 2019.

\bibitem{Zhong2019Invariance_cvpr}
Z.~Zhong, L.~Zheng, Z.~Luo, S.~Li, and Y.~Yang, ``Invariance matters: Exemplar
  memory for domain adaptive person re-identification,'' in \emph{{IEEE}
  Conference on Computer Vision and Pattern Recognition}, 2019, pp. 598--607.

\bibitem{Wu2019Unsupervised_cvpr}
J.~Wu, Y.~Yang, H.~Liu, S.~Liao, Z.~Lei, and S.~Z. Li, ``Unsupervised graph
  association for person re-identification,'' in \emph{{IEEE} Conference on
  Computer Vision and Pattern Recognition}, 2019.

\bibitem{zhong2018generalizing}
Z.~Zhong, L.~Zheng, S.~Li, and Y.~Yang, ``Generalizing a person retrieval model
  hetero- and homogeneously,'' in \emph{Proceedings of the European Conference
  on Computer Vision}, 2018, pp. 176--192.

\bibitem{Wu_2019_CVPR}
A.~Wu, W.~Zheng, X.~Guo, and J.~Lai, ``Distilled person re-identification:
  Towards a more scalable system,'' in \emph{{IEEE} Conference on Computer
  Vision and Pattern Recognition}, 2019, pp. 1187--1196.

\bibitem{Wang2020Smoothing_cvpr}
G.~Wang, J.~Lai, W.~Liang, and G.~Wang, ``Smoothing adversarial domain attack
  and p-memory reconsolidation for cross-domain person re-identification,'' in
  \emph{{IEEE} Conference on Computer Vision and Pattern Recognition}, 2020.

\bibitem{ma2017self-paced}
F.~Ma, D.~Meng, Q.~Xie, Z.~Li, and X.~Dong, ``Self-paced co-training,'' in
  \emph{Proceedings of the International Conference on Machine Learning}, 2017,
  pp. 2275--2284.

\bibitem{zheng2016mars}
L.~Zheng, Z.~Bie, Y.~Sun, J.~Wang, C.~Su, S.~Wang, and Q.~Tian, ``Mars: A video
  benchmark for large-scale person re-identification,'' in \emph{Proceedings of
  the European Conference on Computer Vision}, 2016, pp. 868--884.

\bibitem{Li2020Block_cvpr}
C.~Li, L.~Yuan, J.~Peng, G.~Wang, X.~Liang, L.~Lin, and X.~Chang,
  ``Block-wisely supervised neural architecture search with knowledge
  distillation,'' in \emph{{IEEE} Conference on Computer Vision and Pattern
  Recognition}, 2020.

\end{thebibliography}

\begin{IEEEbiography}[{\includegraphics[width=1in,height=1.25in,clip,keepaspectratio]{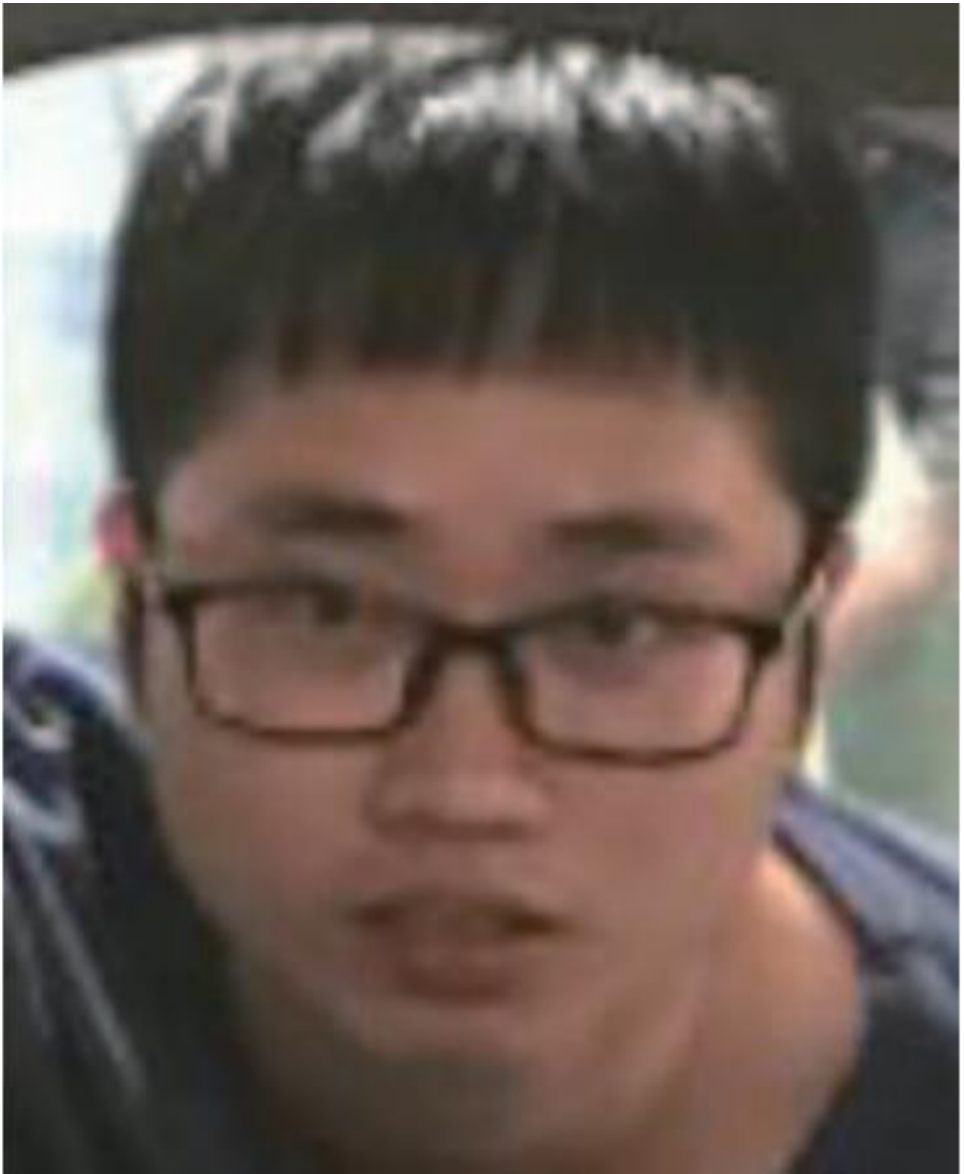}}]{Guangrun Wang} is currently a Ph.D. candidate in the School of Data and Computer Science, Sun Yat-sen University, Guangzhou, China. He received the B.E. degree from Sun Yat-sen University in 2014. From 2015 to 2017, he was a visiting scholar with the Department of Information Engineering, The Chinese University of Hong Kong. His research interests include machine learning, computer vision. He is the recipient of the 2018 Pattern Recognition Best Paper Award and one ESI Highly Cited Paper. 
\end{IEEEbiography}

\begin{IEEEbiography}[{\includegraphics[width=1in,height=1.25in,clip,keepaspectratio]{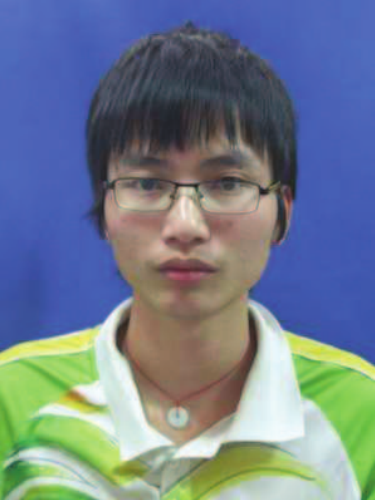}}]{Guangcong Wang} is pursuing a Ph.D. degree in the School of Data and Computer Science, Sun Yat-sen University, Guangzhou, China. He received the B.E. degree in communication engineering from Jilin University (JLU), Changchun, China, in 2015. His research interests are computer vision and machine learning. He has published several works on person re-identification, weakly supervised learning, semi-supervised learning, and deep learning.
\end{IEEEbiography}

\begin{IEEEbiography}[{\includegraphics[width=1in,height=1.25in,clip,keepaspectratio]{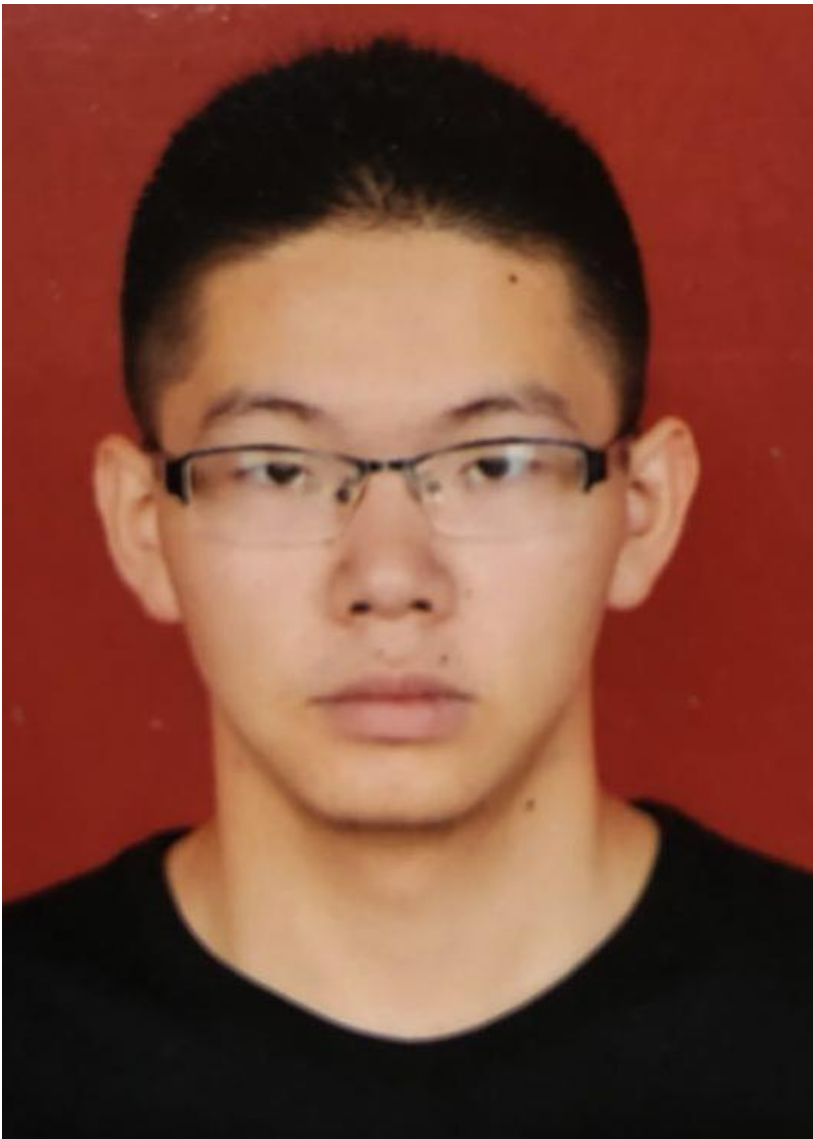}}]{Xujie Zhang} is currently an undergraduate student in the School of Data and Computer Science, Sun Yat-sen University (SYSU), Guangzhou, China. He majors in computer science. His research interest is computer vision and machine learning. Currently, he aims at developing algorithms for person re-identification, especially weakly supervised person re-identification.
\end{IEEEbiography}

\begin{IEEEbiography}[{\includegraphics[width=1in,height=1.25in,clip,keepaspectratio]{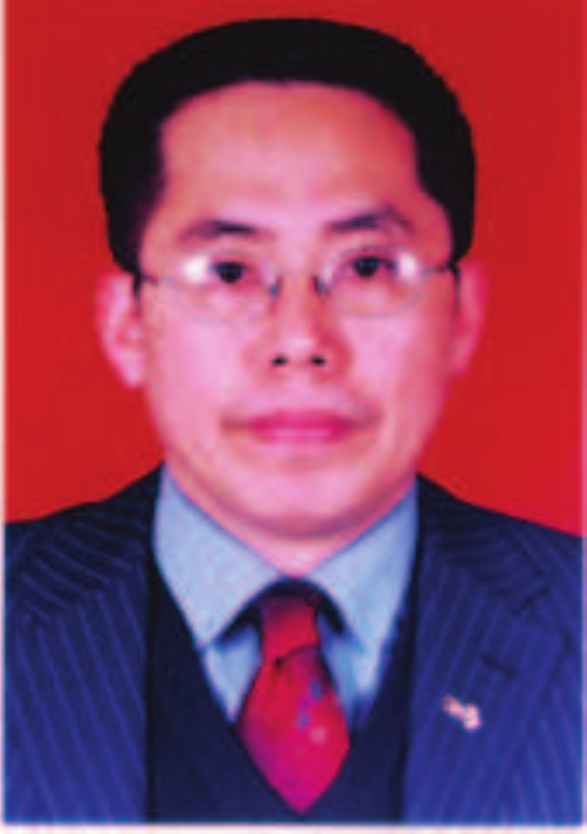}}]{Jianhuang Lai}
received the Ph.D. degree in mathematics in 1999 from Sun Yat-sen University, China. He joined Sun Yat-sen University in 1989 as an assistant professor, where he is currently a Professor of the School of Data and Computer Science. His current research interests are in the areas of digital image processing, pattern recognition, and applications. He has published over 200 scientific papers in academic journals and conferences on image processing and pattern recognition. 
\end{IEEEbiography}

\begin{IEEEbiography}[{\includegraphics[width=1in,height=1.25in,clip,keepaspectratio]{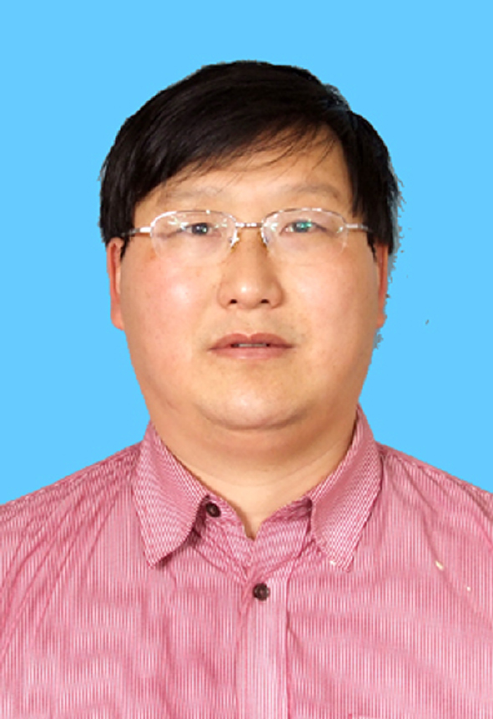}}]{Zhengtao Yu} received the Ph.D. degree in computer application technology from the Beijing Institute of Technology, Beijing, China, in 2005. He is currently a Professor with the School of Information Engineering and Automation, Kunming University of Science and Technology, China. His main research interests include natural language process, image processing, and machine learning.
\end{IEEEbiography}

\begin{IEEEbiography}[{\includegraphics[width=1in,height=1.25in,clip,keepaspectratio]{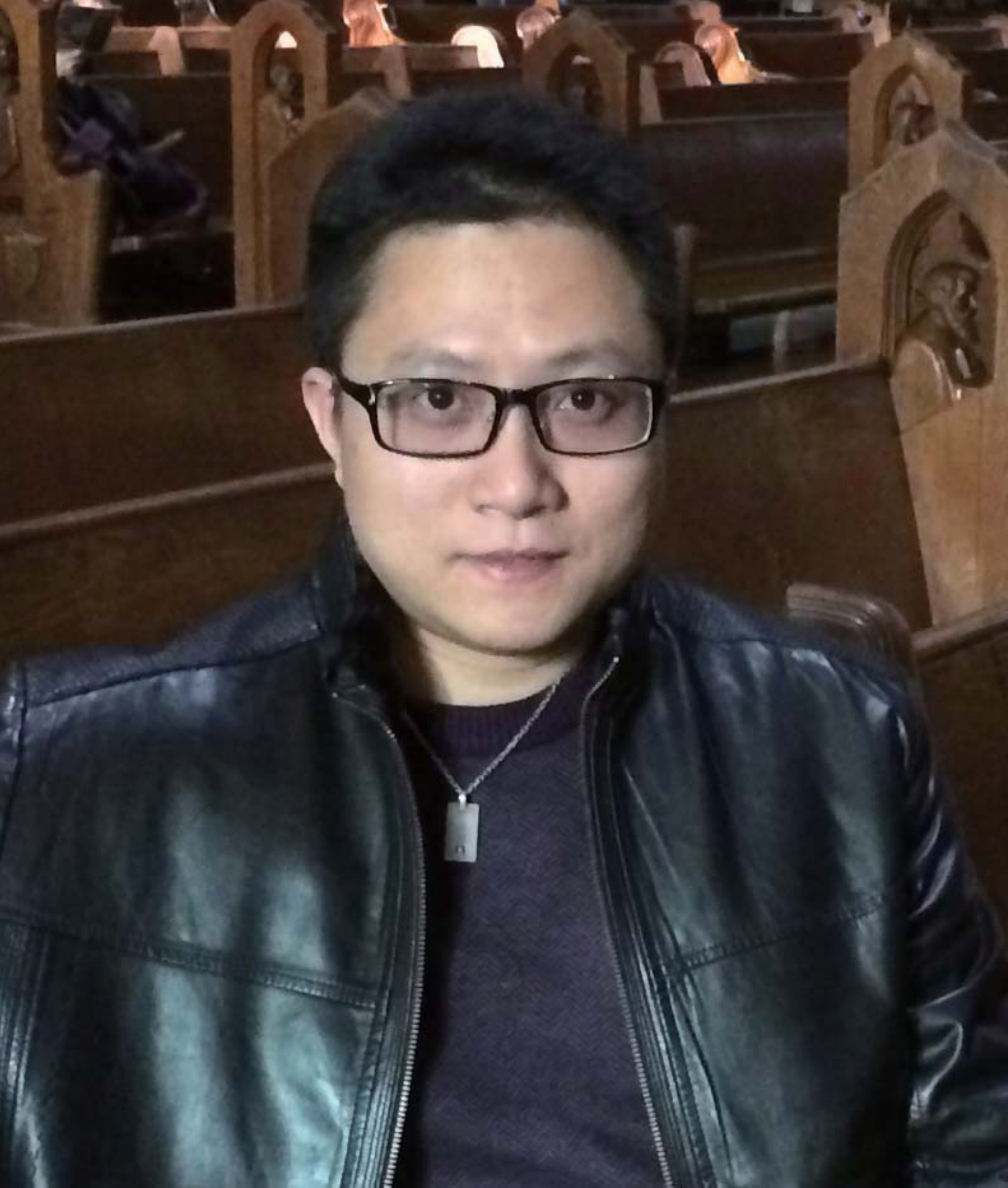}}]{Liang Lin} (M'09, SM'15) a full Professor of Sun Yat-sen University. From 2008 to 2010, he was a Post-Doctoral Fellow at the University of California, Los Angeles. He has authorized and co-authorized on more than 100 papers in top-tier academic journals and conferences. He has been serving as an associate editor of IEEE Trans. Human-Machine Systems. He served as Area Chairs for numerous conferences such as CVPR, and ICCV. He was the recipient of the Google Faculty Award in 2012, Best Paper Diamond Award in IEEE ICME 2017, and Hong Kong Scholars Award in 2014, and the Annual Pattern Recognition Best Paper Award. He is a Fellow of IET.

\end{IEEEbiography}

\ifCLASSOPTIONcaptionsoff
  \newpage
\fi
\end{document}